\documentclass[10pt,twocolumn,letterpaper]{article}

\usepackage[pagenumbers]{cvpr} 

\usepackage{multirow} 

\definecolor{cvprblue}{rgb}{0.21,0.49,0.74}
\usepackage[pagebackref,breaklinks,colorlinks,allcolors=cvprblue]{hyperref}
\usepackage{amsmath}
\usepackage{amssymb} 
\usepackage{caption}
\usepackage{subcaption}
\usepackage{multirow}
\usepackage{graphicx}
\usepackage{caption}
\usepackage{booktabs} 
\usepackage{textcomp}
\usepackage{hyperref}

\renewcommand\footnotemark{}  
\renewcommand\thanks[1]{\footnotemark\protected@xdef\@thanks{\@thanks  
        \protect\footnotetext[0]{#1}}}  
\makeatother  

\title{SkillMimic: Learning Basketball Interaction Skills from Demonstrations}

\author{
Yinhuai Wang$^{1,2*}$ \quad Qihan Zhao$^{1*}$ \quad Runyi Yu$^{1,2*}$ \quad Hok Wai Tsui$^{1}$ \quad Ailing Zeng$^{6\dagger}$  \quad Jing Lin$^{5}$  \\
Zhengyi Luo$^{7}$ \quad Jiwen Yu$^{3}$ \quad Xiu Li$^{4}$ \quad Qifeng Chen$^{1}$ \quad Jian Zhang$^{3\dagger}$ \quad Lei Zhang$^{5}$ \quad Ping Tan$^{1}$ 
\\
\normalsize $^{1}$Hong Kong University of Science and Technology \quad 
\normalsize $^{2}$Unitree Robotics \quad
\normalsize $^{3}$ Peking University Shenzhen Graduate School \\
\normalsize $^{4}$Tsinghua University \quad
\normalsize $^{5}$International Digital Economy Academy \quad \normalsize $^{6}$Tencent \quad \normalsize $^{7}$Carnegie Mellon University
\\
{\tt\small yinhuai.wang@connect.ust.hk} \quad {\tt\small pingtan@ust.hk}
}

\begin{document}

\twocolumn[{
\renewcommand\twocolumn[1][]{#1}
\vspace{-0.3cm}
\maketitle
\centering
\vspace{-0.4cm}
\includegraphics[width=\textwidth]{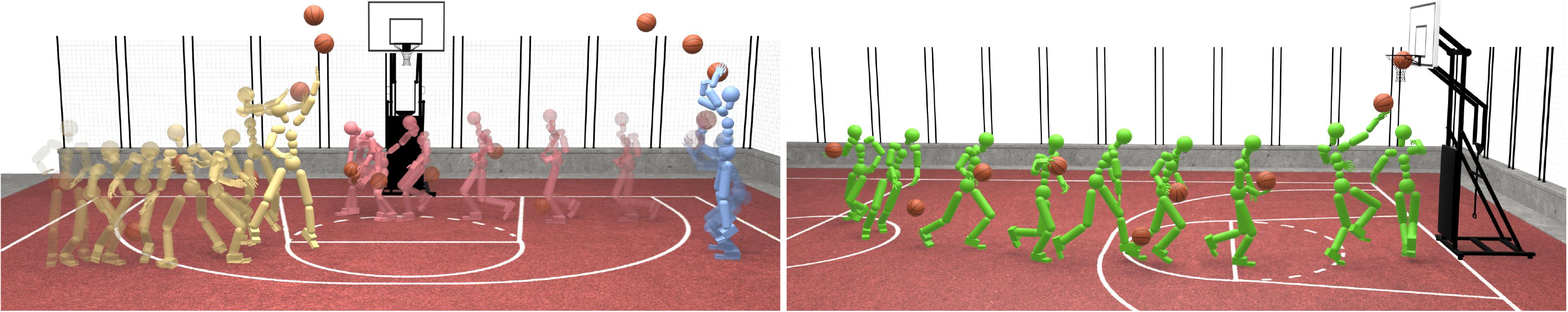}
\vspace{-0.5cm}
\captionsetup{type=figure}
\caption{We propose a novel approach that for the first time enables physically simulated humanoids to learn a variety of basketball interaction skills from Human-Object Interaction (HOI) data, including but not limited to shooting (blue), retrieving (red), and turnaround layup (yellow). Once acquired, these interaction skills can be composed to accomplish complex tasks, such as consecutive scoring (green). 
}
\label{fig: teaser}
\vspace{0.5cm}
}]

\renewcommand{\thefootnote}{\fnsymbol{footnote}}
\footnotetext[0]{$^{*}$Equal Contribution. $^{\dagger}$Corresponding author.
}
\begin{abstract}
Traditional reinforcement learning methods for human-object interaction (HOI) rely on labor-intensive, manually designed skill rewards that do not generalize well across different interactions. We introduce SkillMimic, a unified data-driven framework that fundamentally changes how agents learn interaction skills by eliminating the need for skill-specific rewards. Our key insight is that a unified HOI imitation reward can effectively capture the essence of diverse interaction patterns from HOI datasets. This enables SkillMimic to learn a single policy that not only masters multiple interaction skills but also facilitates skill transitions, with both diversity and generalization improving as the HOI dataset grows.
For evaluation, we collect and introduce two basketball datasets containing approximately 35 minutes of diverse basketball skills. Extensive experiments show that SkillMimic successfully masters a wide range of basketball skills including stylistic variations in dribbling, layup, and shooting. Moreover, these learned skills can be effectively composed by a high-level controller to accomplish complex and long-horizon tasks such as consecutive scoring, opening new possibilities for scalable and generalizable interaction skill learning. 
Project page: \href{https://ingrid789.github.io/SkillMimic/}{https://ingrid789.github.io/SkillMimic/}
\end{abstract}    
\vspace{-0.2cm}
\section{Introduction}
Human-Object Interaction (HOI) represents one of the most fundamental yet challenging capabilities in robotics and character animation \cite{heomnih2o,fuhumanplus,xu2023interdiff,li2023controllable,luo2024omnigrasp,xiaounified,gao2024coohoi}. Among various HOI scenarios, basketball serves as an exemplary testbed that demands mastery of complex, dynamic, and precise interactions.
Despite significant progress, existing physics-based character animation methods predominantly focus on imitation learning of locomotion skills \cite{DeepMimic,amp,ase,dou2023c,tessler2023calm} through Reinforcement Learning (RL), leaving a significant gap for HOI imitation. While some methods can learn interaction skills, such as striking pillars \cite{ase}, playing tennis \cite{tennis}, climbing ropes \cite{bae2023pmp}, and carrying boxes \cite{hassan2023synthesizing}, they typically involve handcrafted reward functions for each different interaction skill. 
However, designing skill-specific rewards is labor-intensive and fundamentally limits generalization. This limitation becomes particularly acute in basketball, where the complexity and diversity of skills would demand extraordinary engineering effort for reward design. Consequently, existing methods \cite{DeepMimic,amp,ase,dou2023c,tessler2023calm,tennis,bae2023pmp,hassan2023synthesizing} fail to provide a unified learning framework for a single policy to learn diverse interaction skills, let alone achieving complex long-term tasks such as consecutive basketball scoring.

How can a simulated humanoid learn a wide variety of interaction skills in a simple and scalable manner? Consider how humans naturally acquire complex skills: amateur basketball players often develop their expertise by watching and imitating game footage, even without formal coaching. This self-directed learning process involves synchronizing their body movements with both the observed player motions and ball trajectories through practice.

Inspired by these observations, we propose a data-driven method called SkillMimic, which mimics HOI through RL to learn interaction skills. SkillMimic can learn various interaction skills purely from HOI datasets, such as diverse styles of basketball shooting, layup, and dribbling skills, and even a robust pickup skill that enables picking up balls in random locations and motions. Notably, SkillMimic uses the exact same configuration to learn different interaction skills, with identical hyperparameters. This allows us to train a single Interaction Skill (IS) policy to learn multiple interaction skills and achieve smooth skill switching. By training a high-level policy to reuse these learned interaction skills, we can accomplish challenging high-level tasks such as scoring, which requires the humanoid to composite diverse basketball interaction skills to score accurately.

To advance basketball skill learning, we introduce two HOI datasets: BallPlay-V, which captures player and ball motion from RGB videos across eight basic skills, and BallPlay-M, which uses optical motion capture to record 35 minutes of comprehensive basketball interactions. Using these datasets, SkillMimic successfully learns diverse basketball skills through a unified IS policy. With an additional high-level controller for skill composition, our system can execute complex tasks like consecutive scoring. Our approach demonstrates superior efficiency in learning both interaction skills and multi-phase, long-horizon tasks compared to existing methods, opening new possibilities for scalable and generalizable interaction skill learning.

Specifically, our contributions are as follows:
\begin{itemize}
\item \textbf{SkillMimic}: a data-driven paradigm for learning diverse humanoid interaction skills through RL. It supports joint learning and switching of diverse basketball interaction skills, with skill diversity and generalization improving as the demonstration data grows, showcasing the scalability.
\item \textbf{Contact graph}: We propose a simple and general contact modeling method that applies to diverse skills, called the contact graph. A contact graph reward is designed to enable precise contact imitation, which proves to be critical for learning precise interaction skills.
\item \textbf{Unified HOI imitation reward}: We propose a set of important designs that form a unified reward configuration for imitation learning of various interaction skills. 
\item \textbf{A hierarchical solution for reusing interaction skills}: We propose training a high-level controller to reuse the interaction skills acquired by SkillMimic to accomplish challenging high-level tasks, which for the first time achieves continuous basketball scoring.
\item \textbf{BallPlay Datasets}: We introduce two basketball datasets to facilitate research on interaction skill learning.
\end{itemize}

\section{Related Work}

\paragraph{Locomotion Imitation Learning.}
Mimicking human motion for robot control is an efficient method of learning humanoid skills. DeepMimic \cite{DeepMimic}, a pioneer in this field, uses imitation learning to perform a variety of highly dynamic locomotion skills. The introduction of Generative Adversarial Imitation Learning (GAIL) \cite{ho2016generative} into humanoid imitation learning by AMP \cite{amp} lessens constraints on data alignment, thus enhancing its versatility. ASE \cite{ase} further amplifies motion diversity in GAIL training, incorporates a pre-trained low-level policy to acquire locomotion skills, and deploys a high-level policy to repurpose these locomotion skills for specific tasks, such as striking pillars. These methodologies are subsequently expanded in numerous efforts to augment condition and text control \cite{juravsky2022padl,tessler2023calm,dou2023c,zhu2023neural,sun2023prompt,ren2024insactor,pan2024synthesizing,luo2024smplolympics,won2022physics}, as well as to design automated rewards \cite{cui2024anyskill}. 
However, a limitation of these studies is their exclusive focus on learning isolated locomotion skills through imitation. This requires additional pipeline and reward design for learning interaction skills, even for relatively simple ones, such as moving boxes \cite{hassan2023synthesizing} or striking pillars \cite{ase}. 
In contrast, our approach enables unified learning of diverse interaction skills, eliminating the need for skill-specific reward engineering.

\vspace{-0.3cm}
\paragraph{Learning Human-Object Interaction (HOI).}
Learning HOI in simulation has long posed challenges in the fields of graphics and robotics \cite{li2023object,li2023controllable,xu2023interdiff,xu2024interdreamer,jiang2024scaling,jiang2023full,Starke2020Local,Starke2021Neural,wu2022saga,ghosh2022imos}. Early studies often relied on manually designed control structures. For example, Hodgins et al. \cite{hodgins1995animating} implement a variety of sports interactions through the use of manually designed state machine controllers. Yin et al. \cite{yin2007simbicon} use finite-state machines to generate a large variety of gaits. Coros et al. \cite{coros2010generalized} utilize an inverted pendulum model to control foot placement. Recently, deep reinforcement learning \cite{arulkumaran2017deep} has been widely employed in character control, giving rise to an array of methods based on policy network control. These methods have been applied to a range of challenging activities, such as basketball \cite{liu2018learning}, skateboarding \cite{liu2017learning}, and cycling \cite{tan2014learning}, etc. The use of imitation rewards further lessens the reward design for locomotion styles. Zhang et al. \cite{tennis} propose a framework to learn diverse tennis skills from broadcast videos. Bae et al. \cite{bae2023pmp} incorporate multiple part-wise motion priors for diverse whole-body interactions. Similarly, Braun et al. \cite{braun2023physically} develop a framework for synthesizing whole-body grasps.

Applying imitation learning for HOI is intuitive, however, effectively learning diverse interaction skills in a unified framework remains unresolved. Naively extending locomotion imitation methods \cite{DeepMimic,amp} to HOI imitation often yields unstable and unnatural results, as these methods do not address the unbalanced rewards inherent in HOI, nor do they properly account for the crucial aspects of relative motion and contact in HOI. Modeling interaction through interaction graphs is a well-established approach \cite{ho2010spatial}. While this method has proven useful for HOI imitation \cite{zhang2023simulation}, it remains limited by considering only kinematic relationships, failing to ensure proper contact and becoming unstable in learning diverse scenarios. In contrast, our contact graph imitation reward explicitly prioritize physical contact learning, significantly improving the success rate of HOI imitation and enabling unified HOI imitation of diverse complex interaction skills for the first time. 

This work builds upon and significantly extends ideas initially explored in our early technical report \cite{wang2023physhoi}.

\begin{figure}[t]
  \centering  \includegraphics[width=\linewidth]{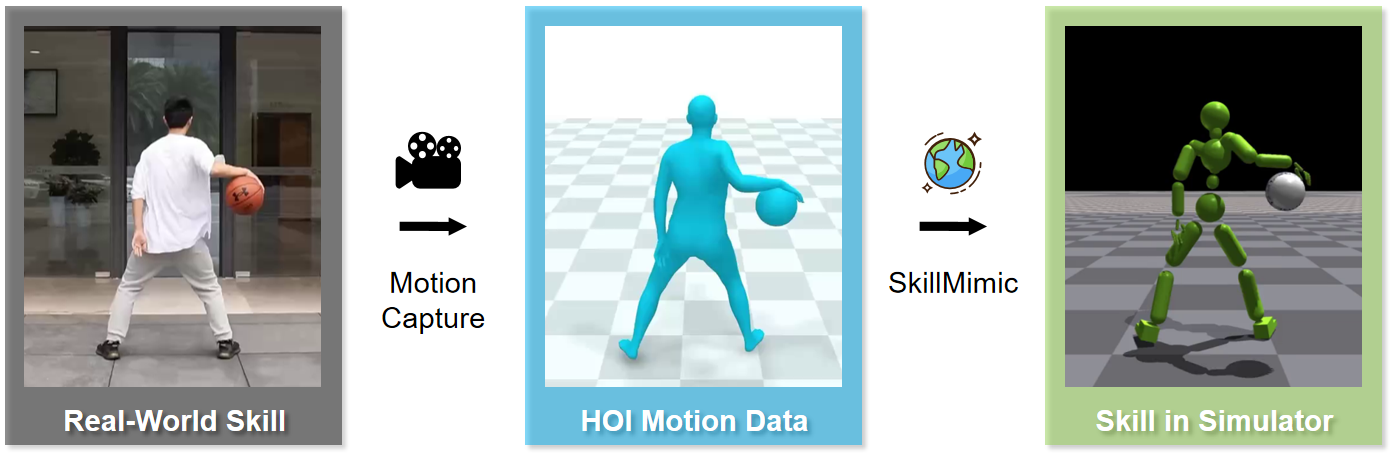}
  \caption{Concept of SkillMimic. We define an interaction skill as a set of Human-Object Interaction (HOI) state transitions that align with the intended skill semantics. These state transitions can be derived from captured HOI motion clips. If a simulated humanoid can manipulate objects such that the resulting HOI state transitions closely match those of the reference, we consider the humanoid to have successfully learned the interaction skill.} 
\label{fig: skill imitation}
\end{figure}

\begin{figure*}[t]
  \centering  \includegraphics[width=0.9\linewidth]{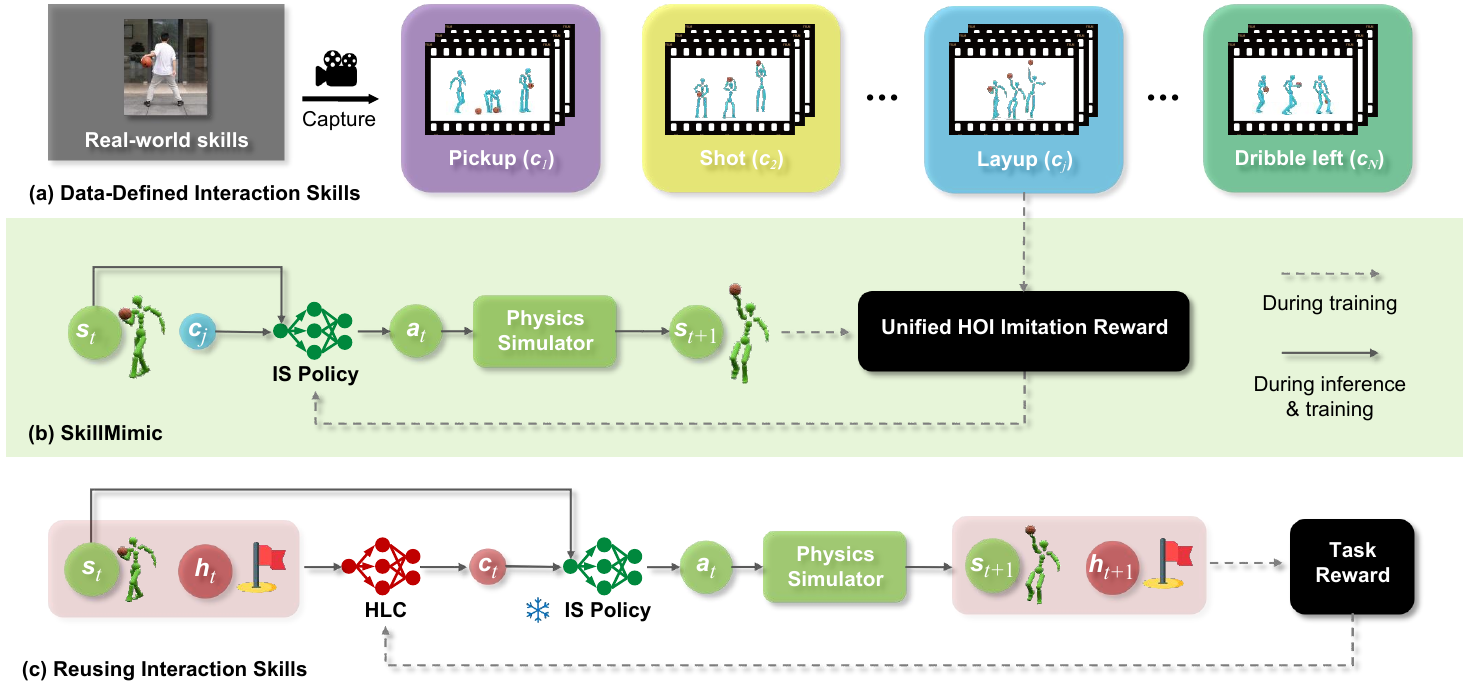}
  \caption{Our system consists of three parts. (a) First, we capture real-world basketball skills to create a large Human-Object Interaction (HOI) motion dataset. (b) Second, we train an Interaction Skill (IS) policy to learn interaction skills by imitating the corresponding HOI data through reinforcement learning. Specifically, the IS policy takes as input the HOI state $\boldsymbol{s}_{t}$ and skill label $\boldsymbol{c}_{j}$ and predicts the action $\boldsymbol{a}_{t}$. The new state $\boldsymbol{s}_{t+1}$ is calculated by the simulator. A unified HOI imitation reward is designed to imitate diverse HOI state transitions. (c) The third part involves training a High-Level Controller (HLC) to reuse the learned interaction skills for complex tasks. The HLC takes as input $\boldsymbol{s}_{t}$ and extra task observations $\boldsymbol{h}_{t}$, e.g., the basket position, and predicts the skill label $\boldsymbol{c}_{t}$ to drive a pre-trained IS policy.} 
\label{fig: overview}
\end{figure*}

\section{SkillMimic}
In this section, we introduce a unified data-driven framework for simulated humanoids to learn various basketball interaction skills via simulation and reinforcement learning (RL). 
We begin by introducing how we define interaction skills with Human-Object Interaction (HOI) data (Sec.~\ref{sec: define skills}). Next, we explain how to learn diverse interaction skills by imitating HOI data (Sec.~\ref{sec: train skills}) and discuss the designing of a unified HOI imitation reward (Sec.~\ref{sec: unified imitation reward}). Finally, we demonstrate how the learned interaction skills can be reused to perform high-level tasks (Sec.~\ref{sec: reuse skills}).

\subsection{The BallPlay Dataset}
\label{sec: ballplay}
To address the scarcity of basketball HOI data and facilitate research on interaction skill learning, we introduce two datasets: BallPlay-V, based on monocular vision estimation, containing eight clips of different basketball skills; and BallPlay-M, based on multi-view optical motion capture systems, containing 35 minutes of diverse basketball interactions. We provide more details in the appendix.

\subsection{Defining Interaction Skills Using HOI Data} 
\label{sec: define skills}

We propose a data-driven approach to defining interaction skills as collections of Human-Object Interaction (HOI) state transitions that align with the intended skill semantics. For instance, the interaction skill of ``picking up balls from various positions" can be encapsulated by various HOI motion clips, each capturing a unique instance of the action. These clips collectively form a collection of HOI state transitions that represent the interaction skill. 

\subsection{Learning Interaction Skills by HOI Imitation} 
\label{sec: train skills}
Considering an interaction skill defined by a set of reference HOI state transitions, if a humanoid can manipulate objects such that its HOI state transitions closely resemble the reference, we consider the humanoid to have successfully learned the interaction skill. Based on this concept, we propose a method for learning interaction skills by imitating HOI state transitions through RL, which we call SkillMimic. Fig.~\ref{fig: skill imitation} illustrates the concept of SkillMimic. 

\begin{figure*}[t]
  \centering
   \begin{subfigure}[b]{0.4\columnwidth}
    \includegraphics[width=\linewidth]{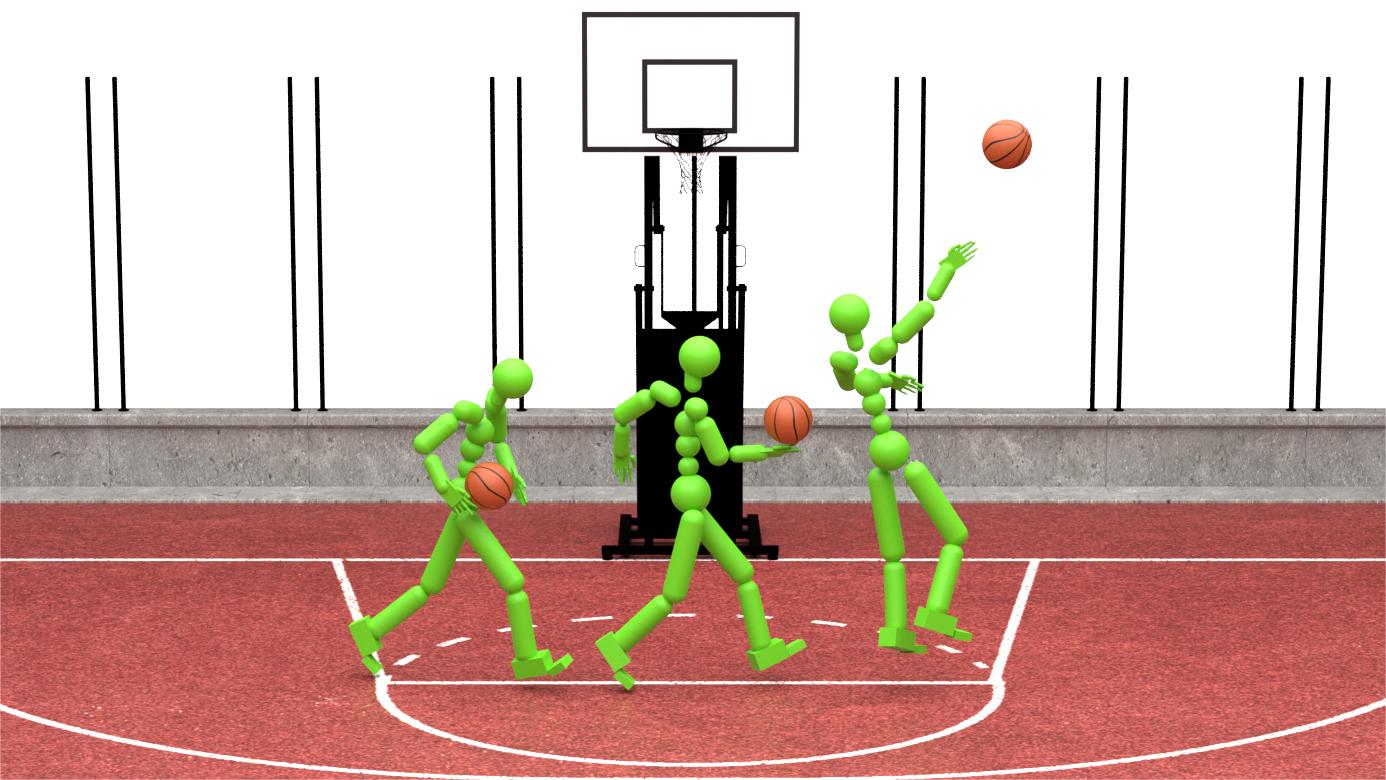}
    \caption{Layup}
  \end{subfigure}
  \begin{subfigure}[b]{0.4\columnwidth}
    \includegraphics[width=\linewidth]{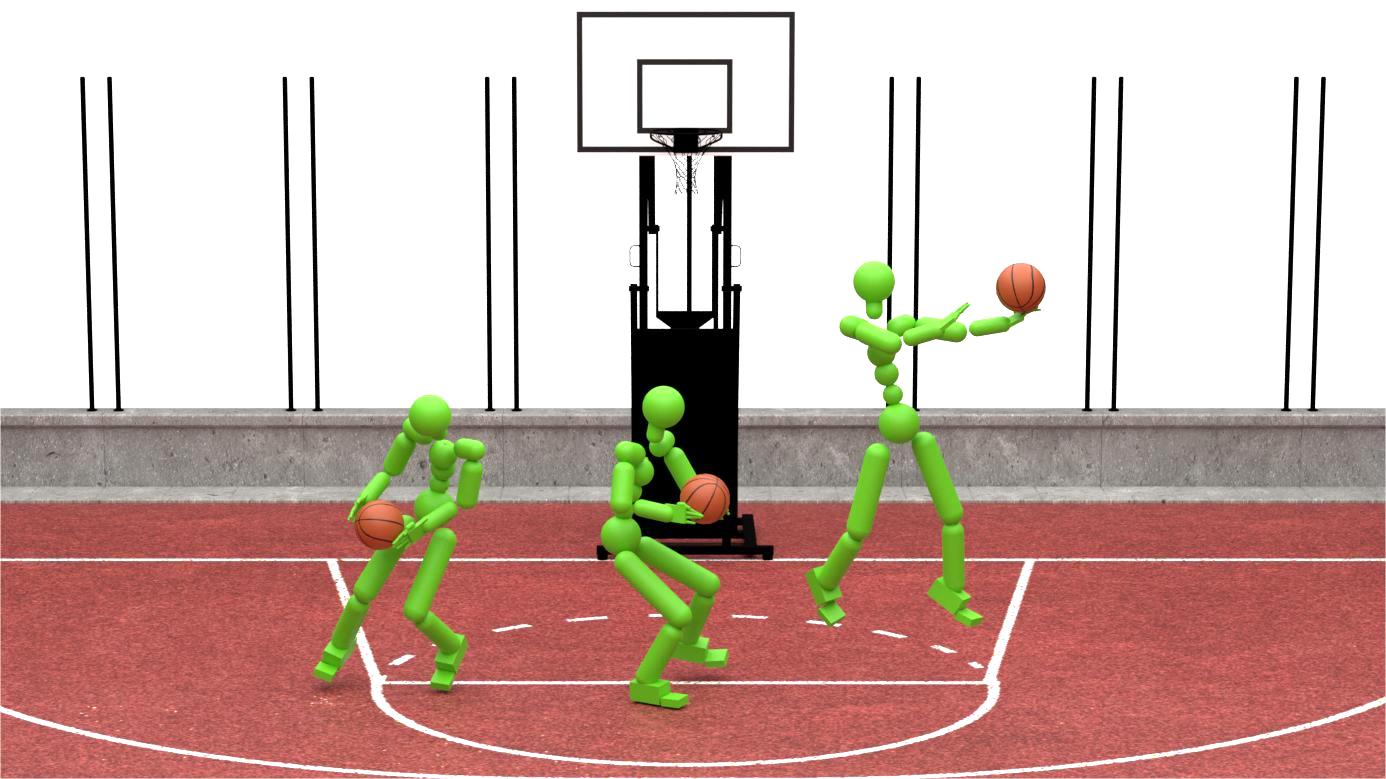}
    \caption{Turnaround Layup}
  \end{subfigure}
  \begin{subfigure}[b]{0.4\columnwidth}
    \includegraphics[width=\linewidth]{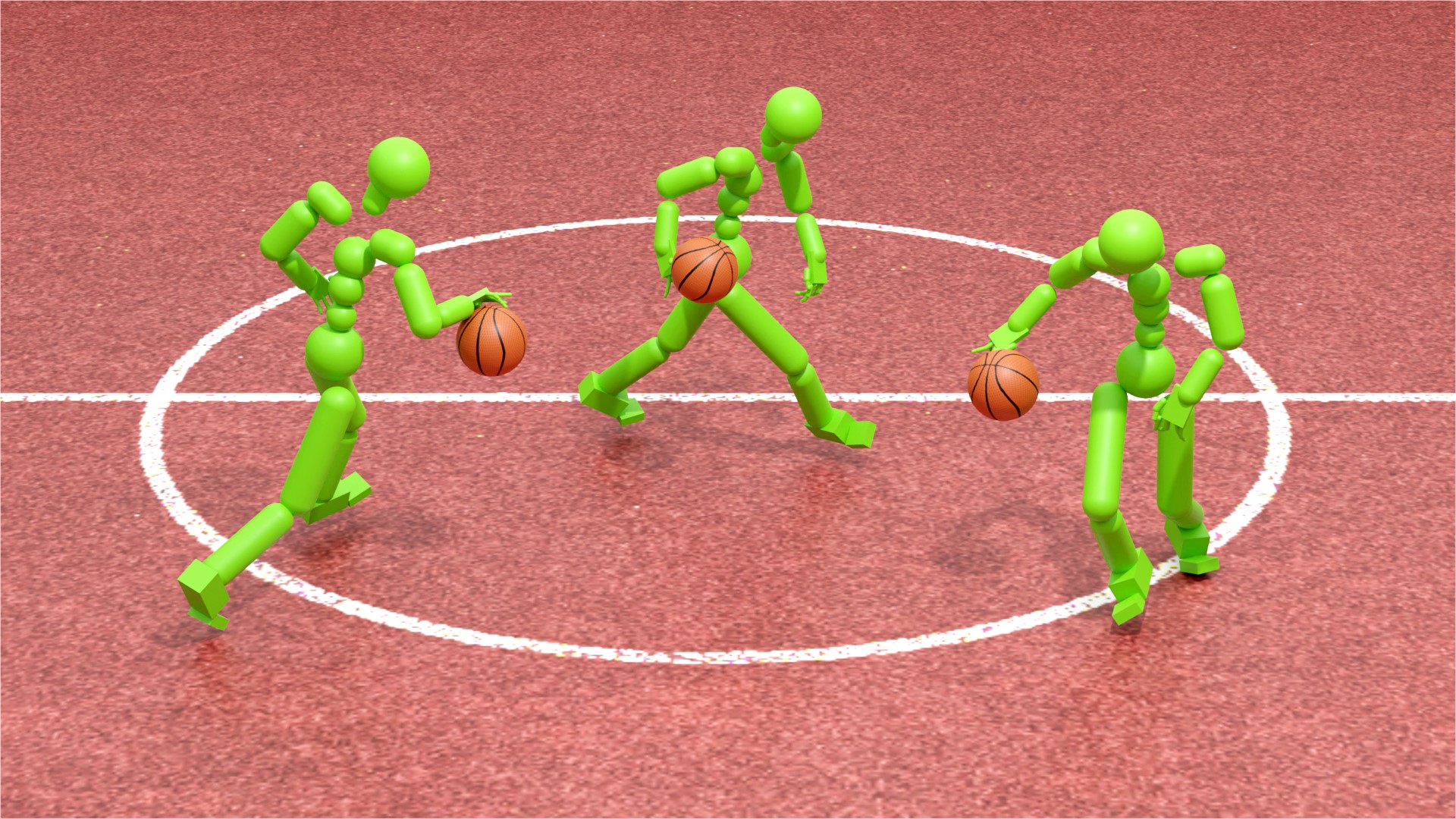}
    \caption{Dribble Right}
  \end{subfigure}
  \begin{subfigure}[b]{0.4\columnwidth}
    \includegraphics[width=\linewidth]{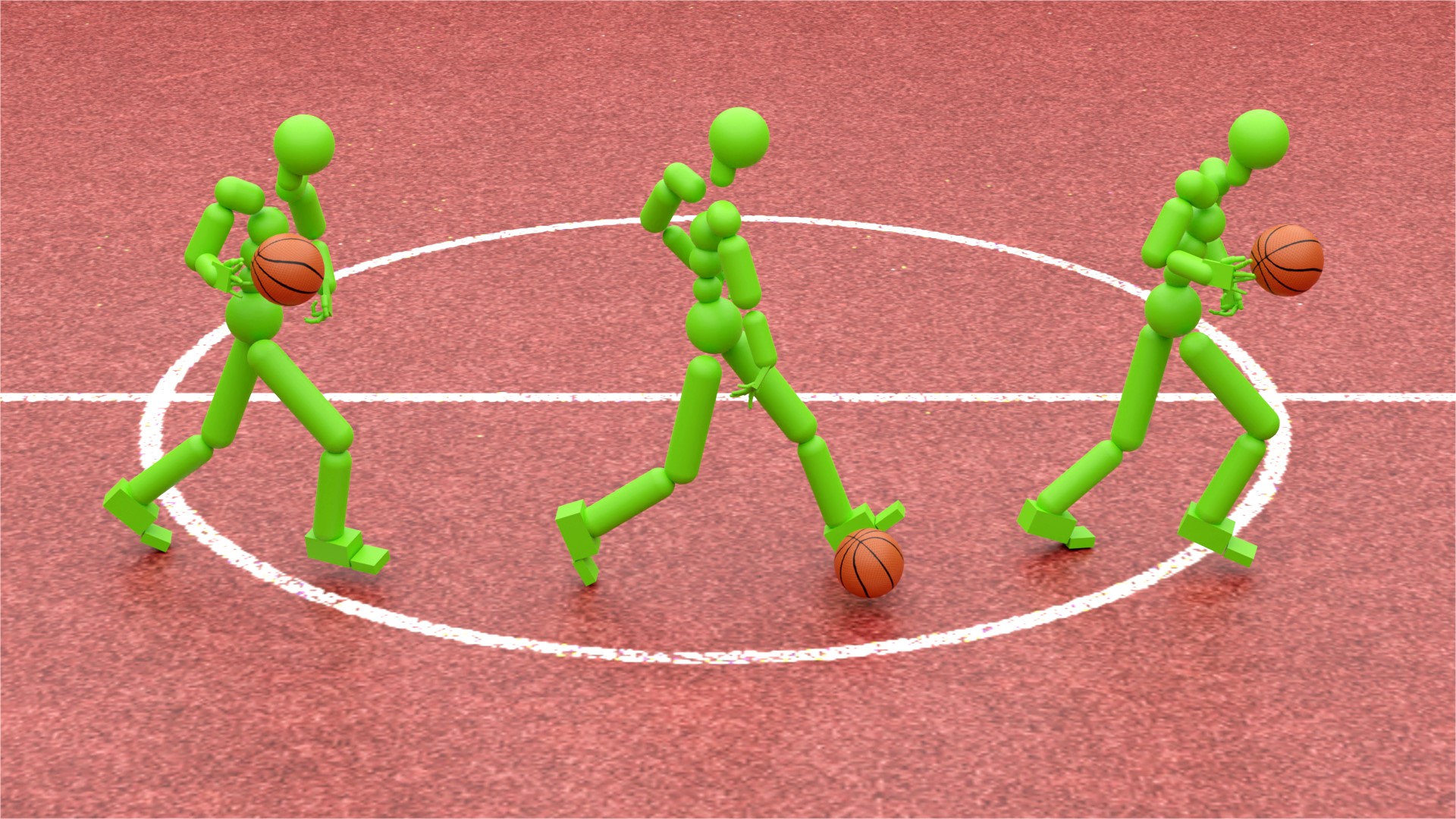}
    \caption{Dribble Forward}
  \end{subfigure}
  \begin{subfigure}[b]{0.4\columnwidth}
    \includegraphics[width=\linewidth]{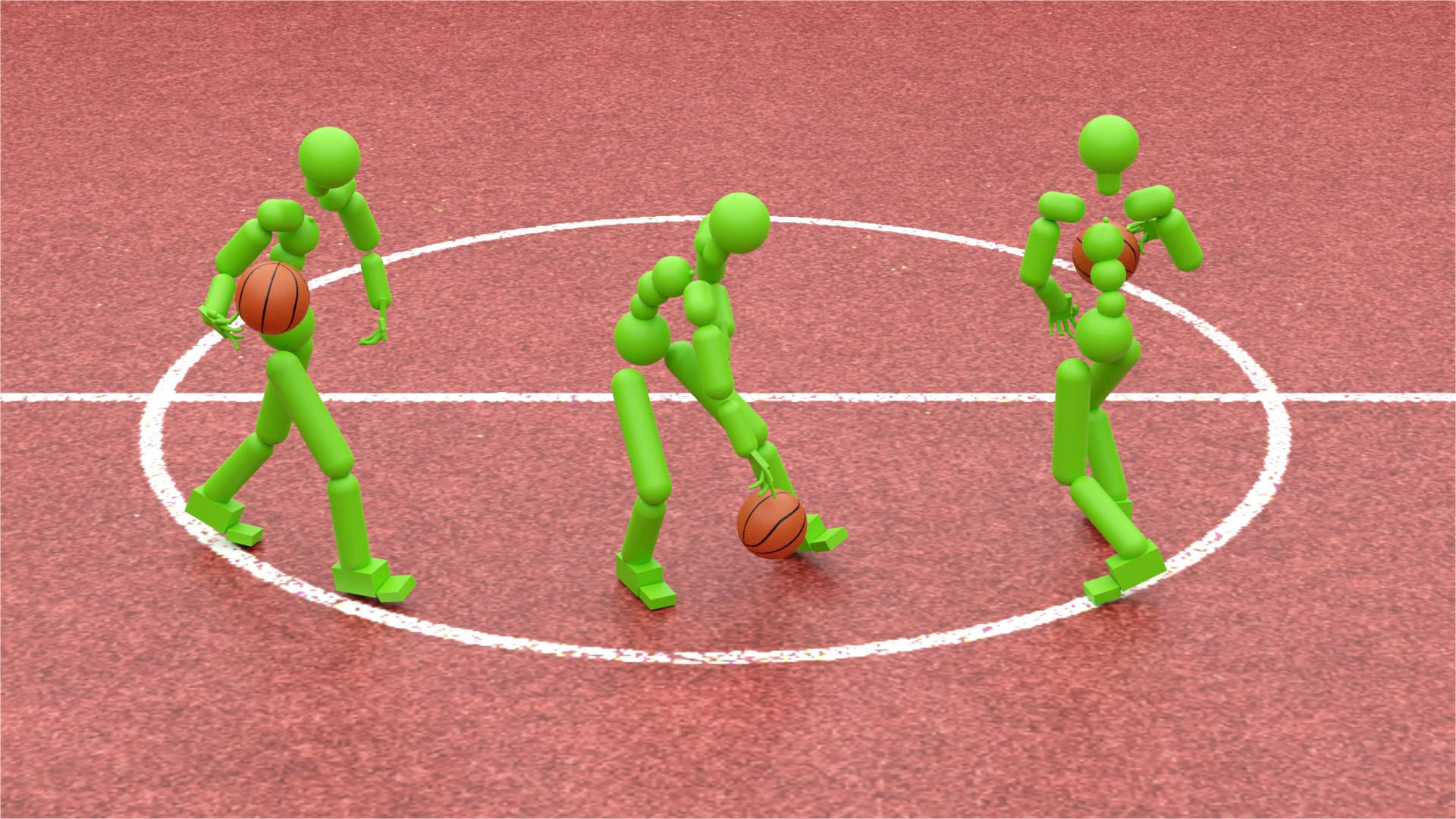}
    \caption{Dribble Turn}
  \end{subfigure}
  \begin{subfigure}[b]{0.4\columnwidth}
    \includegraphics[width=\linewidth]{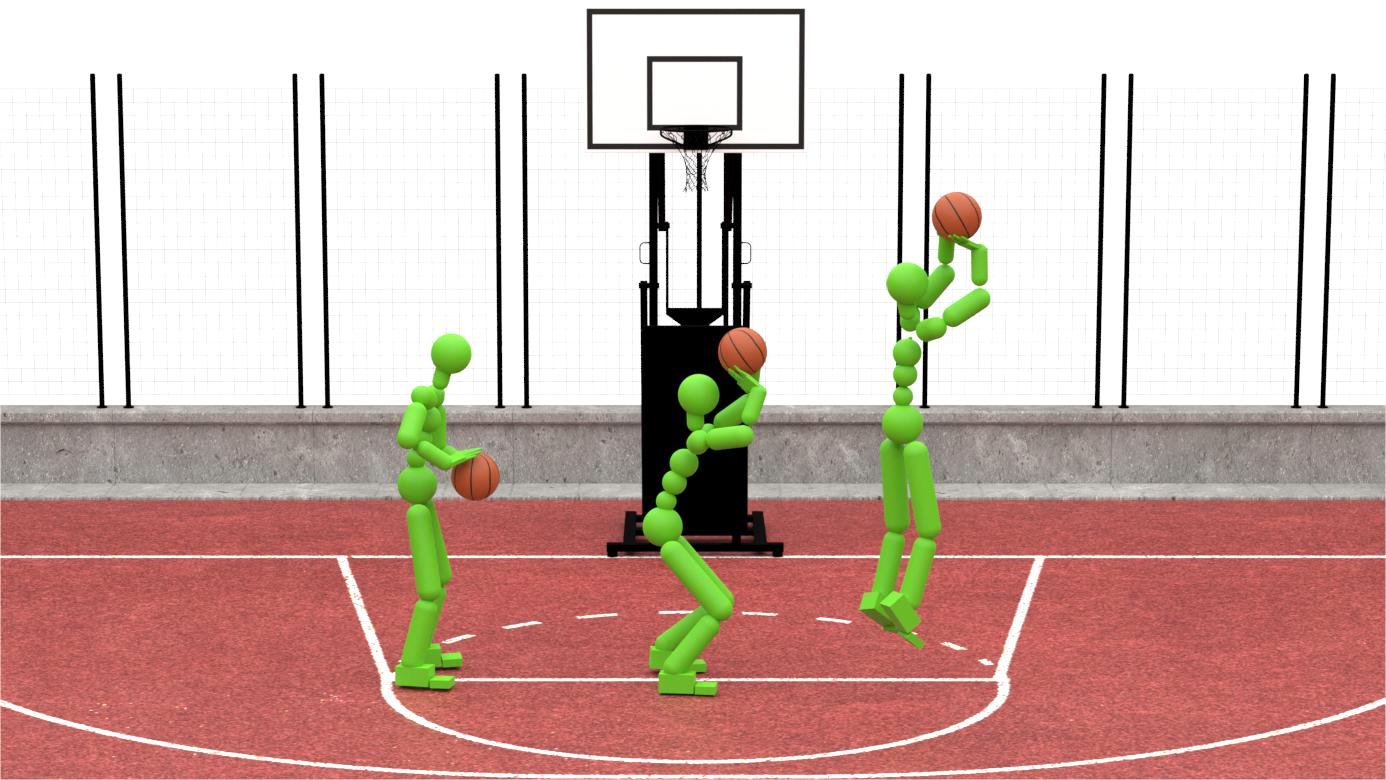}
    \caption{Jump Shot}
  \end{subfigure}
  \begin{subfigure}[b]{0.4\columnwidth}
    \includegraphics[width=\linewidth]{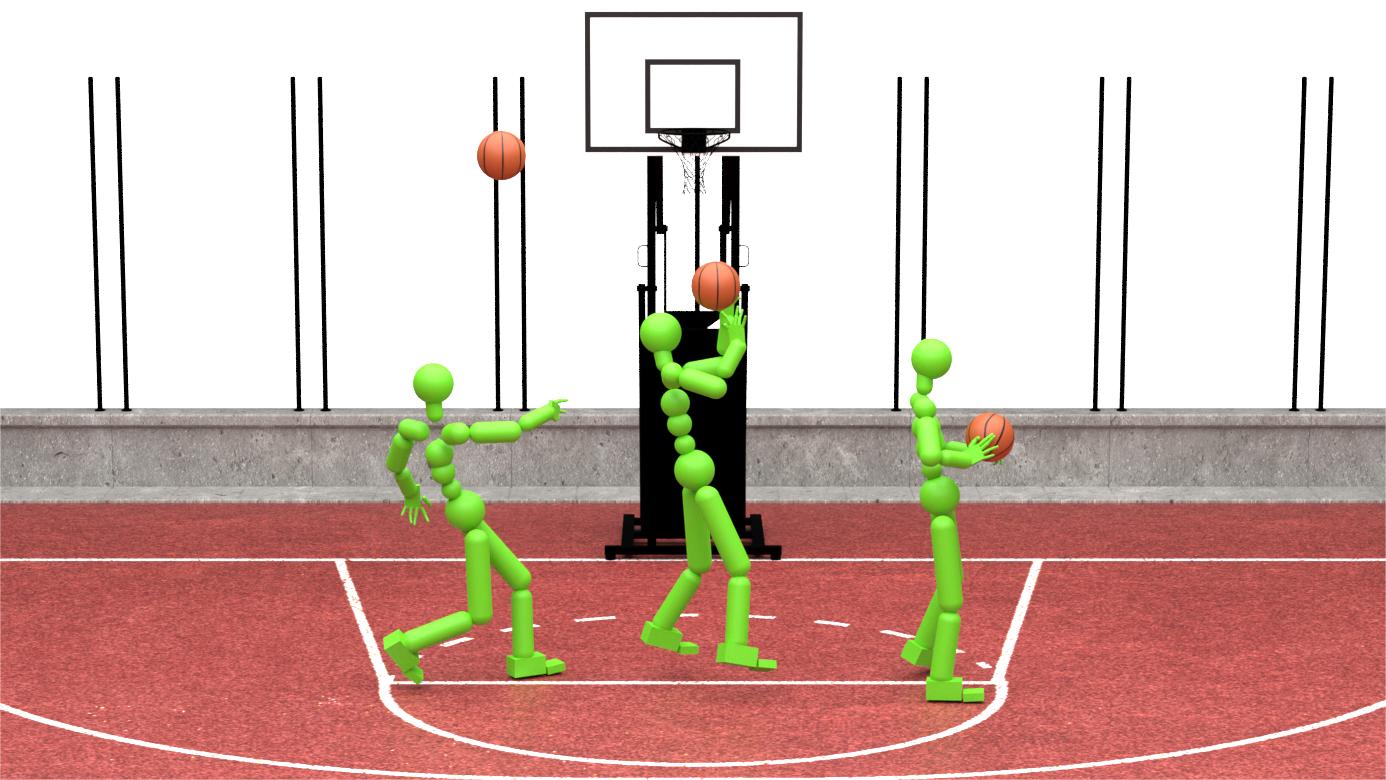}
    \caption{Rebound}
  \end{subfigure}
  \begin{subfigure}[b]{0.4\columnwidth}
    \includegraphics[width=\linewidth]{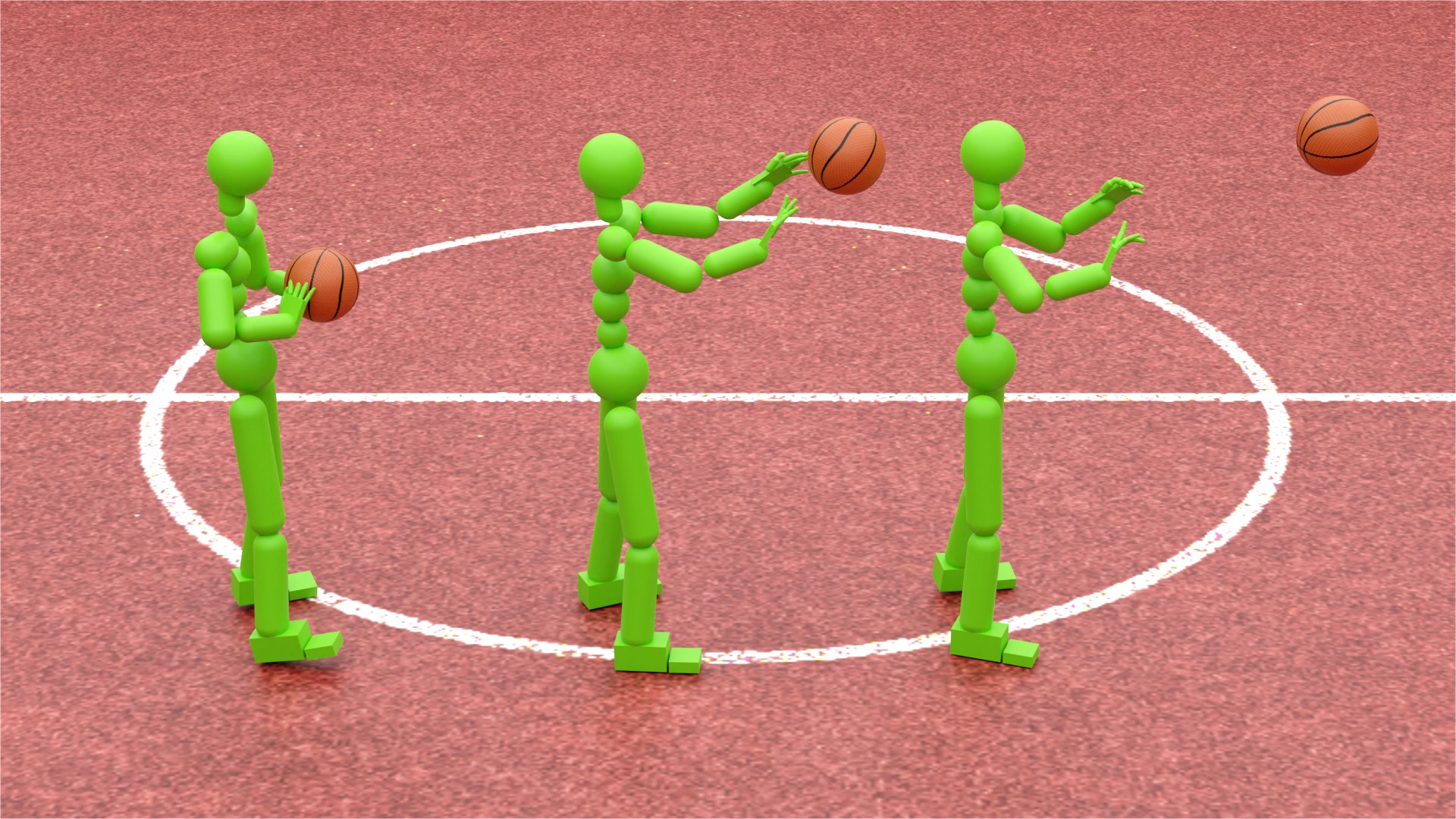}
    \caption{Pass}
  \end{subfigure}
  \begin{subfigure}[b]{0.4\columnwidth}
    \includegraphics[width=\linewidth]{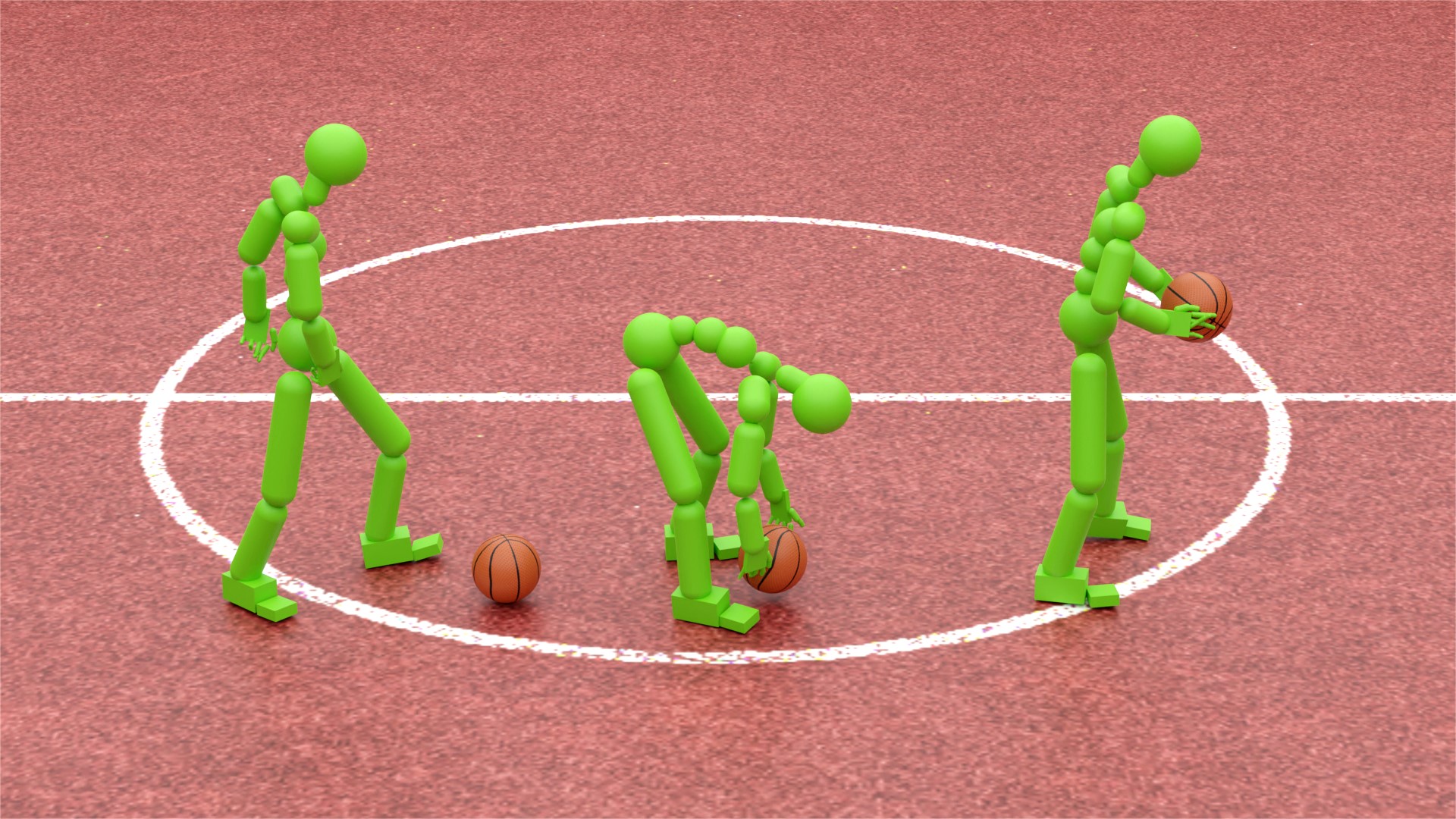}
    \caption{Pickup}
  \end{subfigure}
  \begin{subfigure}[b]{0.4\columnwidth}
    \includegraphics[width=\linewidth]{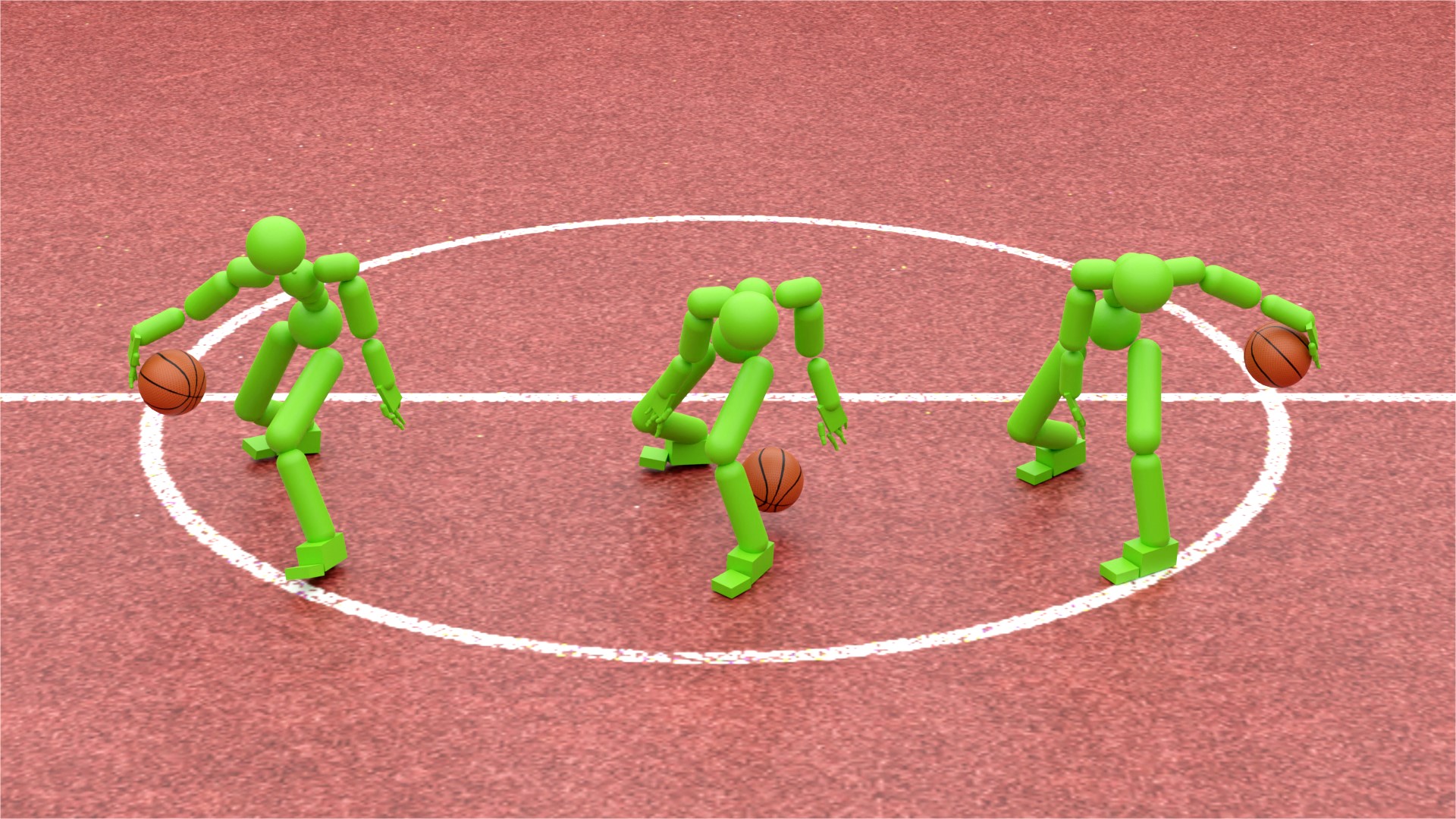}
    \caption{Cross Leg}
  \end{subfigure}
    \caption{Simulated humanoids exhibit comprehensive basketball skills. SkillMimic can teach humanoids a wide range of basketball skills using the same configuration in a purely data-driven manner, covering almost all fundamental basketball skills. Keyframes are placed in chronological order from left to right.}
    \label{fig: llc}
\end{figure*}

In contrast to Behavior Clone \cite{black2024pi_0, fu2024mobile, brohan2023rt} methods which rely on precise state-action data, SkillMimic learns actions through RL using state-only trajectories, demonstrating better tolerance to data noise and higher data efficiency. Unlike previous RL methods \cite{amp,ase,tennis} that require manually designed rewards for each interaction skill, SkillMimic is fully data-driven, skill-agnostic, and scalable, making it capable of learning a wide range of basketball skills within a unified solution. Fig.~\ref{fig: llc} shows diverse basketball skills acquired using SkillMimic. Fig.~\ref{fig: skill diversity} illustrates how the performance of the pickup skill improves as the reference data increases.

\subsubsection{Training Pipeline}
Fig.~\ref{fig: overview} (b) shows the training pipeline of SkillMimic. Given an HOI dataset with diverse skill-labeled clips (Fig.~\ref{fig: overview} (a)), SkillMimic trains an Interaction Skill (IS) policy by randomly selecting a frame from a clip corresponding to a chosen skill to initialize the humanoid and object states. The state $\boldsymbol{s}_t$ (Sec.~\ref{sec: obs}) and skill label $\boldsymbol{c}_j$ are input into the IS policy, which predicts actions (Sec.~\ref{sec: action}) that are then simulated to produce the next state $\boldsymbol{s}_{t+1}$. A unified HOI imitation reward (Sec.~\ref{sec: unified imitation reward}) is designed to measure the consistency between the simulated state transitions and the reference transitions. When the humanoid falls or reaches the maximum simulation length, the environment resets, and the process repeats. 
After training converges, the IS policy enables the humanoid to execute interaction skills that closely mirror the reference demonstrations.

\subsubsection{HOI Observation}
\label{sec: obs}
The state observed by the IS policy is not under direct supervision and may theoretically encompass any information that is available in the simulation environment. Importantly, this information does not necessarily have to be included in the reference HOI data.
Similar to prior arts \cite{ase}, we transform all coordinates into the root local coordinate of the humanoid, which aligns the data distribution and benefits the generalization performance. For the humanoid, we observe its global root height, local body position, rotation, position velocity, and angular velocity. These representations form the humanoid proprioception $\boldsymbol{o}^{prop}_{t}$. In addition, we detect net contact forces $\boldsymbol{o}^{f}_{t}$ for all fingertips, which helps to sense contact and accelerate training. For the object, we observe its local position, rotation, velocity, and angular velocity, which form the object observation $\boldsymbol{o}^{obj}_{t}$. Finally, the state perceived by the IS policy is
\begin{equation}
\begin{aligned}
    \boldsymbol{s}_{t} = \{\boldsymbol{o}^{prop}_{t},\boldsymbol{o}^{f}_{t},\boldsymbol{o}^{obj}_{t}\}.
    \label{eq: st}
\end{aligned}
\end{equation}

Skill labels $\boldsymbol{c}_j$ are also used as inputs to the IS policy to differentiate different skills. We use one-hot encodings to represent the skill label.

\subsubsection{IS Policy and Action}
\label{sec: action}
The policy output is modeled as a Gaussian distribution with dimensions equal to the DOF number of the humanoid robot, featuring constant variance. The mean is modeled by a three-layer MLP consisting of [1024, 512, 512] units with ReLU~\cite{nair2010rectified} activations. We use the action $\boldsymbol{a}_{t}$ sampled from the policy as the target joint rotations for a full set of PD controllers. The PD controllers adjust and output the joint torques to reach the target rotations.

\begin{figure}[t]
  \centering  \includegraphics[width=\linewidth]{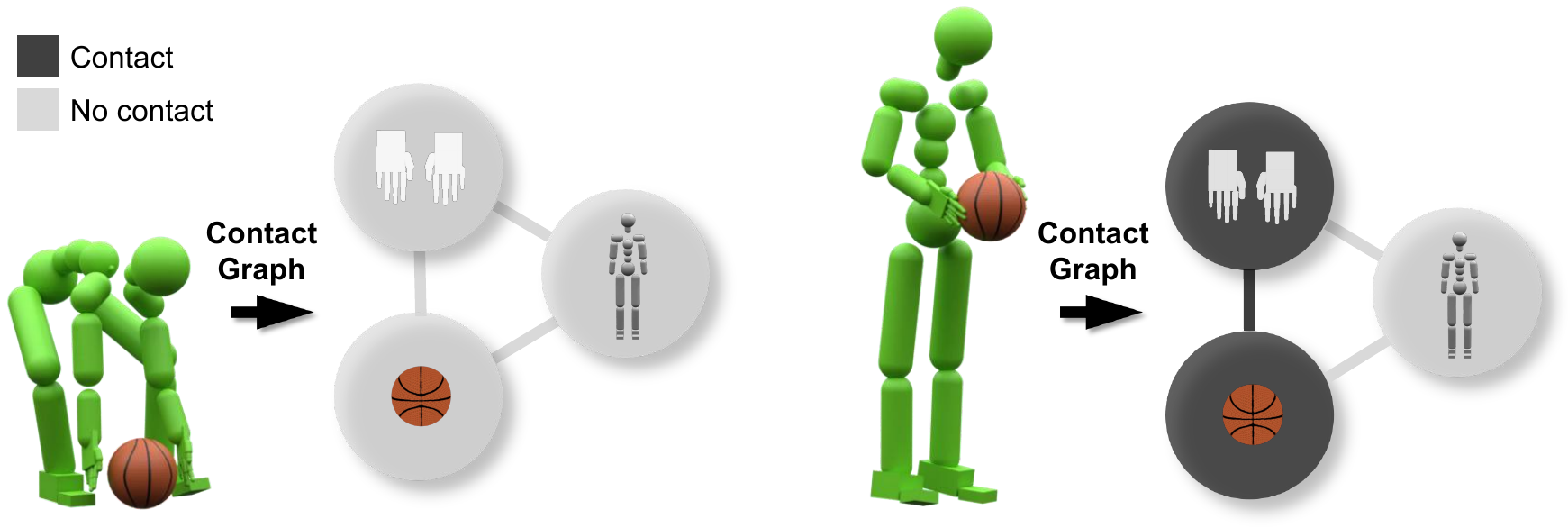}
  \caption{We propose the Contact Graph (CG) to model general contacts within an explicitly defined scene. The node stores a binary value that denotes whether it contacts other nodes. Each edge stores a binary value indicating whether the two connected nodes are in contact. The node definition is unified for a certain scene and shared between diverse interactive skills. For example, we define three nodes: hands, hands-exclusive body, and ball, to form a simple CG to model contacts for diverse basketball skills.} 
\label{fig: cg}
\end{figure}

\subsection{Unified HOI Imitation Reward} 
\label{sec: unified imitation reward}

This HOI imitation reward function is pivotal as it must not only accurately gauge the efficacy of interaction skill acquisition but also avoid the incorporation of skill-specific settings to maintain robust cross-skill generalizability.

Classic imitation rewards often focus solely on imitating locomotions \cite{DeepMimic}, lacking an established baseline for HOI imitation.
Although adversarial imitation rewards \cite{amp,ase} are effective in locomotion tasks, we find that they perform poorly in HOI imitation. This may be because adversarial imitation rewards are relatively coarse-grained, making it difficult for them to provide the precise guidance needed for accurate interaction learning.

The body kinematics imitation reward $r_{t}^{b}$ encourages the alignment of body movements by considering joint position, rotation, and their velocities. Similarly, the object kinematics imitation reward $r_{t}^{o}$ ensures consistent object movements with the reference. However, solely relying on these two rewards often leads to unstable learning and unnatural motions. We incorporate a relative motion reward $r_{t}^{rel}$ to highlight the relative relationships in interactions, which is crucial for interaction naturalness. Recognizing the interlocking relationship between interaction and contact, we propose a straightforward and universal contact modeling approach called Contact Graph (CG). A corresponding CG imitation reward $r_{t}^{cg}$ is designed to enhance the precision of interaction imitation. Additionally, we propose an adaptive velocity regularization term $r_{t}^{reg}$ to suppress high-frequency jitters. These sub-rewards are multiplied rather than added to encourage balanced learning, thereby avoiding local optima, as justified in Tab.~\ref{tab: ablation of HOI imitation}. The combination of these innovations forms a unified imitation reward, enabling the learning of diverse basketball skills in BallPlay-M with the same configuration. To conclude, the complete HOI imitation reward is
\begin{equation}
\begin{aligned}
    r_{t} = r_{t}^{b}*r_{t}^{o}*r_{t}^{rel}*r_{t}^{reg}*r_{t}^{cg}.
    \label{eq: r}
\end{aligned}
\end{equation}

Next, we will discuss the methodology involved in crafting the CG reward $r_{t}^{cg}$. Details about the kinematic imitation reward $r_{t}^{b}*r_{t}^{o}*r_{t}^{rel}*r_{t}^{reg}$ are in the appendix.

\subsubsection{Contact Graph}
\label{sec: Contact Graph}

We observe that existing imitation rewards \cite{amp,ase,zhang2023simulation} only measure kinematic properties and are insufficient to measure precise contacts. To tackle this problem, we propose the Contact Graph (CG) to model the contact in general interactions and design a CG imitation reward.

As illustrated in Fig.~\ref{fig: cg}, the CG is a complete graph where every pair of distinct nodes is connected by a unique edge, defined as $\mathcal{G}=\{\mathcal{V},\mathcal{E}\}$, where $\mathcal{V}\in\{0,1\}^{k}$ is the set of $k$ nodes and $\mathcal{E}\in\{0,1\}^{k(k-1)/2}$ is the set of edges. A CG node stores a binary value indicating whether it contacts other nodes. Each CG edge stores a binary label that denotes the contact between two nodes, where 1 represents contact, and 0 means no contact. CG node and edge values are calculated frame by frame, explicitly describing the mutual contact relationship between CG nodes at different moments. Contacts inside a node are not considered. We use the edge set $\mathcal{E}$ for imitation learning.

The definitions of nodes are flexible. For example, a node can be a single part (e.g., a fingertip), or an aggregation of multiple parts (e.g., the whole left hand). Hinged objects can also be broken up into finer nodes \cite{fan2023arctic,geng2023gapartnet,zhang2024artigrasp}. The definition of CG is unified for a certain scene and shared between diverse interaction skills. For example, in basketball scenarios, we can aggregate two hands as a node, the rest of the bodies as a node, and the ball as a node, as illustrated in Fig.~\ref{fig: cg}. This simple CG is effective enough for learning all the basketball skills covered in BallPlay-M.

\begin{figure}[t]
  \centering  \includegraphics[width=\linewidth]{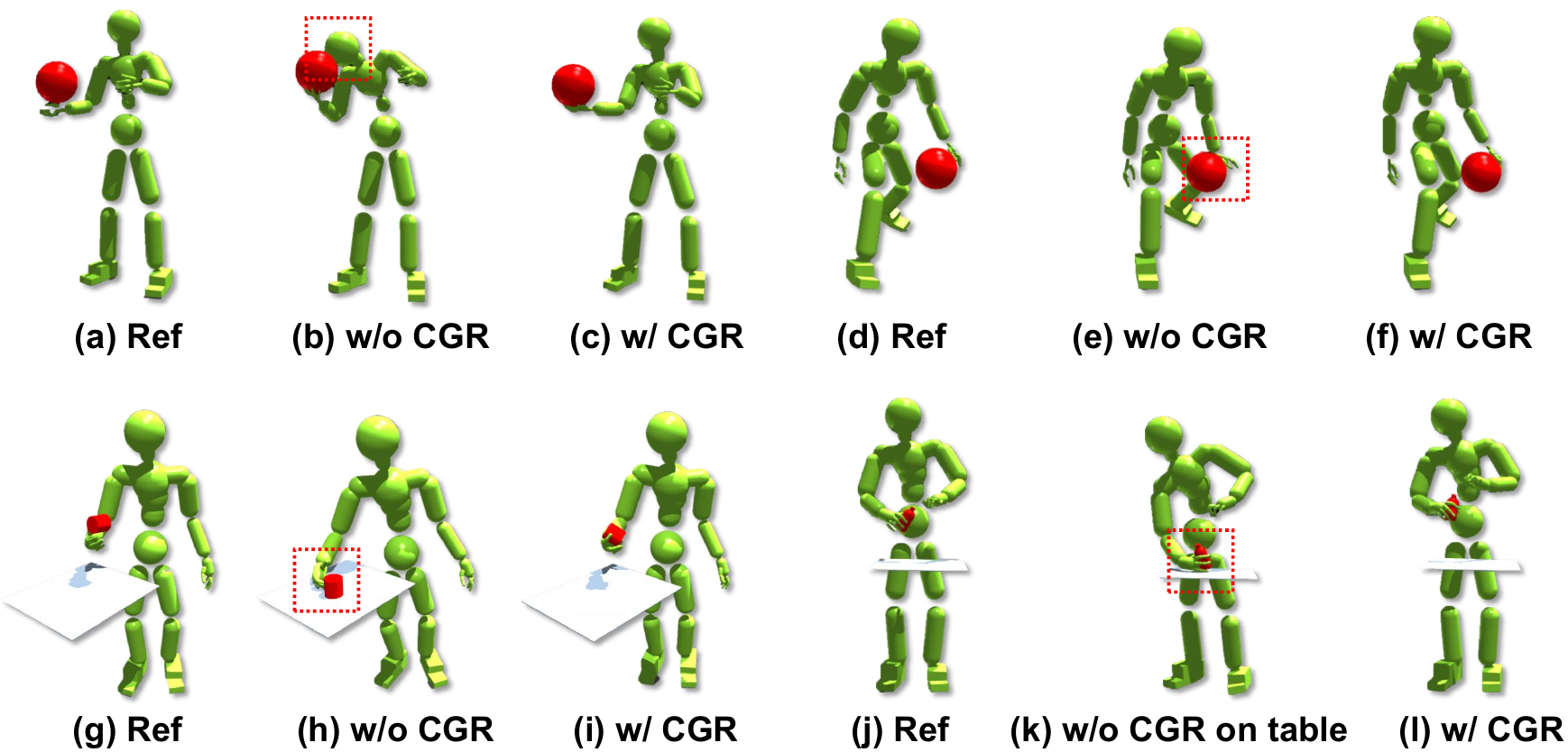}
  \caption{The HOI imitation falls into kinematic local-optimal solutions without Contact Graph Reward (CGR): (\textbf{b}) use the head to help control the ball; (\textbf{e}) use the wrist to contact the ball; (\textbf{h}) fail to catch the object; (\textbf{k}) support the table to keep balance. In comparison, the guidance of CGR effectively yields precise interactions, as shown in (\textbf{c}, \textbf{f}, \textbf{i}, \textbf{l}).} 
\label{fig: cgr}
\end{figure}

\subsubsection{Contact Graph Reward} \label{sec: Contact Graph Reward}
Kinematic imitation rewards \cite{DeepMimic,amp,zhang2023simulation} fall short in measuring contacts and thus yield poor performance in HOI imitation. For example, during the toss skill training (Fig.~\ref{fig: cgr} (a)), the contact between the right hand and the ball usually cause the ball to drop, making the kinematic rewards drastically smaller. In this case, the policy may learn a local-optimal solution that uses the head and left hands to stabilize the ball, as demonstrated in Fig.~\ref{fig: cgr} (b). To tackle this problem, we propose the Contact Graph Reward (CGR) as a critical complement of kinematic imitation rewards to learn precise contact imitation. The CG error is defined as $\boldsymbol{e}_{t}^{cg} = |\boldsymbol{s}_{t}^{cg}-\hat{\boldsymbol{s}}_{t}^{cg}|$,
where $|\cdot|$ calculates element-wise absolute value, $\boldsymbol{s}^{cg}_{t}$ and $\hat{\boldsymbol{s}}^{cg}_{t}$ denotes the simulated and reference CG state respectively, which is the edge set $\mathcal{E}_{t}\in\{0,1\}^{k(k-1)/2}$. The CGR is measured by CG error, with independent weights on different elements:
\begin{equation}
\begin{aligned}
    r_{t}^{cg} = \text{exp}(-\sum_{j=1}^{J}\boldsymbol{\lambda}^{cg}[j] *\boldsymbol{e}_{t}^{cg}[j]), 
    \label{eq: rcg}
\end{aligned}
\end{equation}
where $\boldsymbol{e}_{t}^{cg}[j]$ is the $j$th element of $\boldsymbol{e}_{t}^{cg}\in\{0,1\}^{J}$, representing a contact error; $\boldsymbol{\lambda}^{cg}[j]$ is the $j$th element of $\boldsymbol{\lambda}^{cg}\in\mathbb{R}^{J}$, a hyperparameter controls the sensitivity of a contact. Fig.~\ref{fig: cgr} and Tab.~\ref{tab: ablation of HOI imitation} justify the effectiveness of CGR, where experiments without CGR show inaccurate interactions while using CGR effectively eliminates such problems.

\subsection{Reusing Interaction Skills for High-level Tasks}
\label{sec: reuse skills}

In this section, we propose training a High-Level Controller (HLC) to reuse the interaction skills acquired through SkillMimic to accomplish high-level tasks. As depicted in Fig.~\ref{fig: overview} (c), the HLC takes as input the current HOI state $\boldsymbol{s}_{t}$ and task observation $\boldsymbol{h}_{t}$, and outputs a discrete skill embedding $\boldsymbol{c}_{t}$ that serves as inputs to a pre-trained SkillMimic policy, which subsequently generates actions for the humanoid. 

\begin{figure}[ht]
  \centering
  \begin{subfigure}[b]{0.24\columnwidth}
    \includegraphics[width=\linewidth]{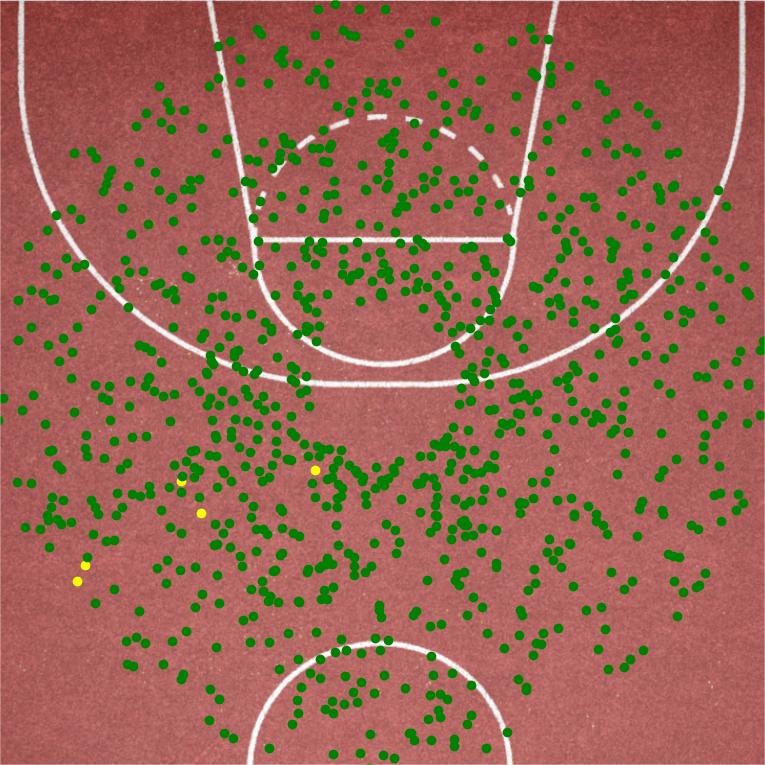}
    \caption{1-0.5\%}
  \end{subfigure}
  \begin{subfigure}[b]{0.24\columnwidth}
    \includegraphics[width=\linewidth]{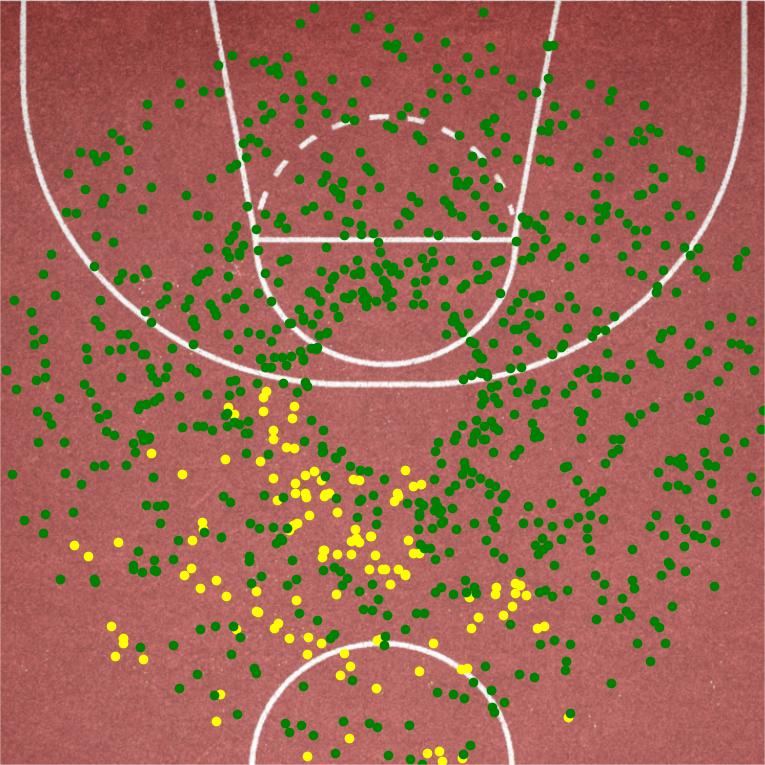}
    \caption{10-12.2\%}
  \end{subfigure}
  \begin{subfigure}[b]{0.24\columnwidth}
    \includegraphics[width=\linewidth]{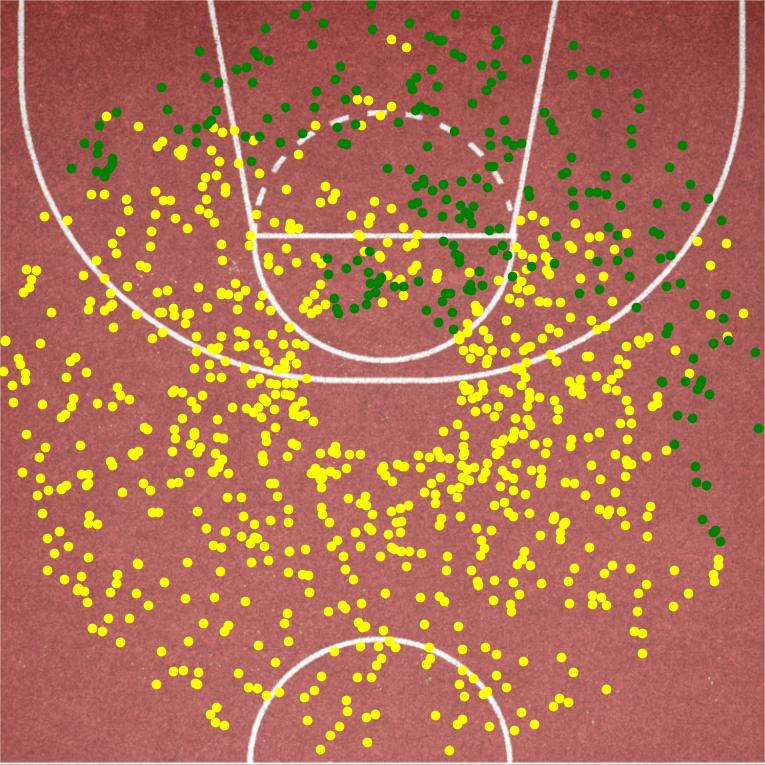}
    \caption{40-76.5\%}
  \end{subfigure}
  \begin{subfigure}[b]{0.24\columnwidth}
    \includegraphics[width=\linewidth]{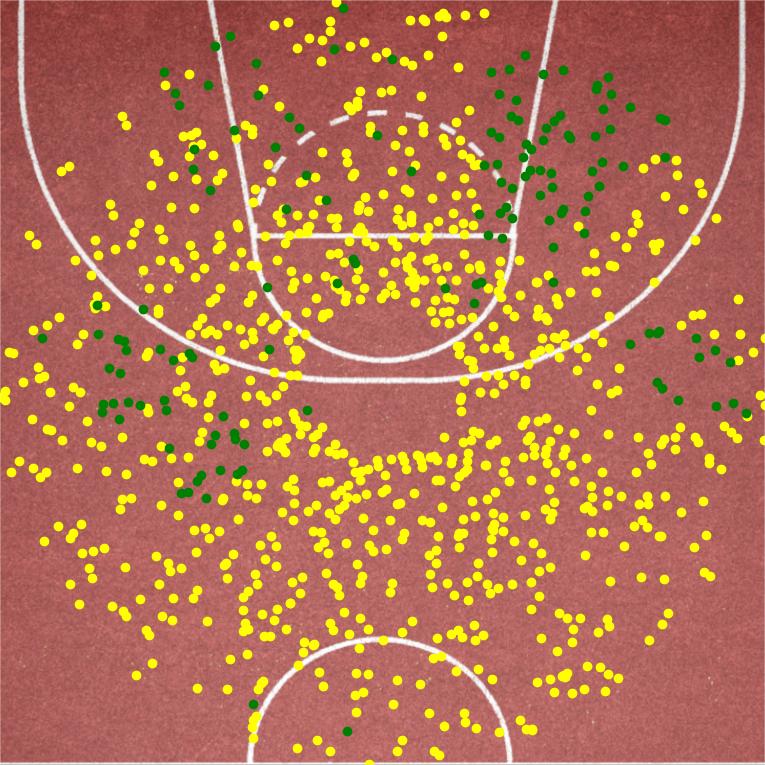}
    \caption{131-85.6\% }
  \end{subfigure}
  \caption{Pickup generalization performance with different training data scales. First number: clips; second: success rate. In each test, 1000 balls are randomly placed within 1 to 5 meters away from the center. Yellow dots indicate successful pickups and green dots represent failures. See Sec.~\ref{ablation:data_scale} for details.}
\label{fig: skill generalization}
\end{figure}

Compared to previous hierarchical approaches that are based on locomotion priors \cite{ase, tennis}, our method leverages interaction skills as prior, which significantly simplifies the design of task rewards and greatly accelerates training. Moreover, unlike their mandatory hierarchical structure, our approach is motivated by data efficiency - tasks can be alternatively solved through direct HOI imitation when sufficient demonstrations are available, as shown in pickup experiments in Sec.~\ref{ablation:data_scale}, Fig.~\ref{fig: skill diversity}, and Fig.~\ref{fig: skill generalization}.

We present four representative high-level tasks:

\begin{figure*}[t]
  \centering  \includegraphics[width=\linewidth]{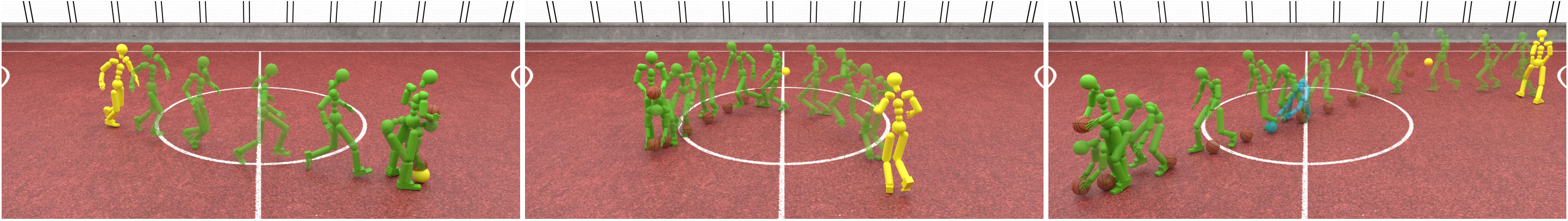}
  \caption{Demonstration of the pickup skill learned from 40 HOI motion clips. Yellow denotes the initial frame. Left: The humanoid picks up a stationary ball effortlessly. Middle: The humanoid intercepts a ball with random velocity. Right: The humanoid adjusts after missing the ball initially (the frame in blue) and successfully retrieves it on the second attempt, showcasing the potential for learning robust and generalizable skills through extensive data collection.} 
\label{fig: skill diversity}
\end{figure*}

\begin{table*}[t]
%
\centering
\resizebox{\linewidth}{!}{%
\begin{tabular}{lr|cccc|cccc|cccc}
\toprule
\multicolumn{2}{c}{} & \multicolumn{4}{c}{SkillMimic w/o Multiplication} & \multicolumn{4}{c}{SkillMimic w/o CGR} & \multicolumn{4}{c}{SkillMimic} 
\\ 
\midrule
Dataset  &  & $\textit{\ Acc.} \uparrow$ & $E_\text{b-mpjpe}  \downarrow$ &  $E_\text{o-mpjpe} \downarrow $ &  $E_\text{cg}\downarrow\ $   & $\ \textit{\ Acc.} \uparrow\ $ & $E_\text{b-mpjpe} \downarrow\ $ & $E_\text{o-mpjpe} \downarrow\ $ &  $E_\text{cg} \downarrow\ $   & $\ \textit{Acc.} \uparrow\ $ & $E_\text{b-mpjpe} \downarrow\ $ &  $E_\text{o-mpjpe} \downarrow\ $ &  $E_\text{cg} \downarrow\ $       \\ \midrule
GRAB \cite{GRAB2020}&  &  {27.0\%} & \textbf{44.7} & {180.2}  & {0.724} &  {38.6\%} & {91.0} & {180.1 }  & {0.337} & \textbf{95.4\%} & {71.1} & \textbf{78.0} & \textbf{0.026} \\
BallPlay-V &  &  {7.5\%} & \textbf{40.1} & {1662.5}  & {0.306} &  {13.6\%} & {88.5} & {155.3} & {0.412} &  \textbf{82.4\%} & {56.8} & \textbf{82.9} & \textbf{0.087} \\
\midrule
\end{tabular}}
    \caption{Ablation study on imitation reward design. We experiment on two HOI datasets: GRAB and BallPlay-V. The result highlights the importance of the CG reward and multiplication in achieving precise HOI imitation.
    }
\label{tab: ablation of HOI imitation}
\end{table*}

\begin{table}[t]
\centering
\resizebox{\linewidth}{!}{%
\begin{tabular}{l|cccc}
\toprule
\multirow{2}{*}{Method} & \multicolumn{4}{c}{Success Rate $\uparrow$} \\ 
\cmidrule(l){2-5}
& Pick Up & Dribble Forward & Layup & Shot \\
\midrule
DeepMimic*\cite{DeepMimic} &  19.6\% & 68.5\%  &  98.9\% & 97.8\% \\
AMP*\cite{amp}  & 0.0\% &  {13.6\%} & 0.0\% & 1.6\% \\
SkillMimic (ours) &   \textbf{86.7\%}  & \textbf{79.6\%}  & \textbf{99.1\%} & \textbf{97.9\%} \\
\midrule
\end{tabular}}
\caption{Success rates across four typical basketball skills in BallPlay-M. Our method significantly outperforms variant methods that use imitation reward styles of DeepMimic and AMP.} 
\label{fig: LLC succ}
\end{table}

\begin{itemize}
\item \textbf{Throwing:} throw the ball to approach a certain height, grab the rebound, and keep on throwing the ball.

\item \textbf{Heading:} dribble the ball to approach the target position.

\item \textbf{Circling:} dribble the ball around the target position following a target radius.

\item \textbf{Scoring:} dribble the ball toward the basket, timing the layup to score, retrieving the rebound, and repeating.
\end{itemize}

Utilizing interaction skills as prior significantly improves learning efficiency for complex, long-horizon tasks. For example, a decent HLC for the Scoring task only requires 3 hours of training on an NVIDIA RTX 4090 GPU.

\begin{figure*}[t]
  \centering
  \vspace{-0.2cm}
  \begin{subfigure}[b]{0.51\columnwidth} \includegraphics[width=\linewidth]{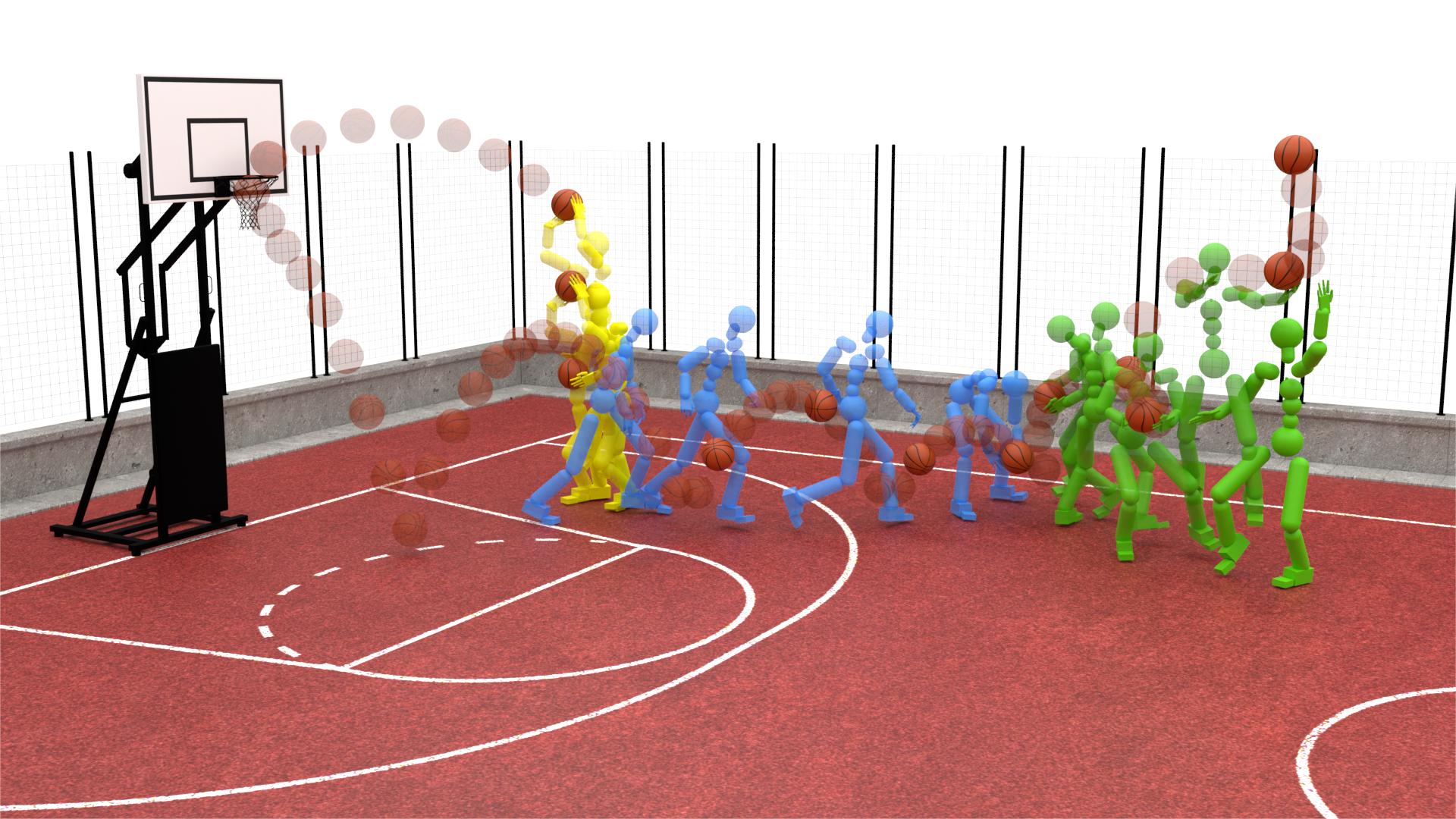}
    \caption{Manual Control}
  \end{subfigure}
  \begin{subfigure}[b]{0.51\columnwidth}
  \includegraphics[width=\linewidth]{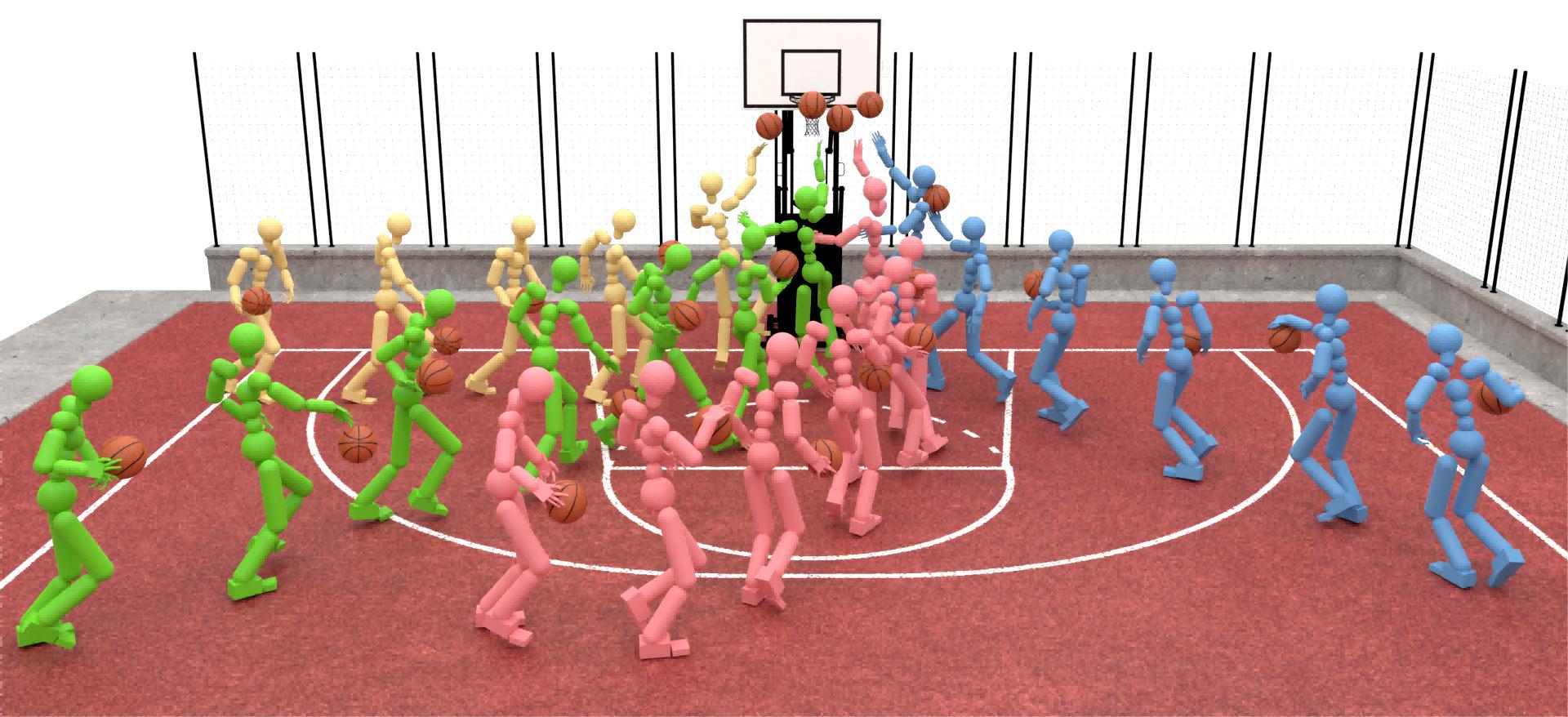}
    \caption{Scoring}
  \end{subfigure}
  \begin{subfigure}[b]
  {0.51\columnwidth}
  \includegraphics[width=\linewidth]{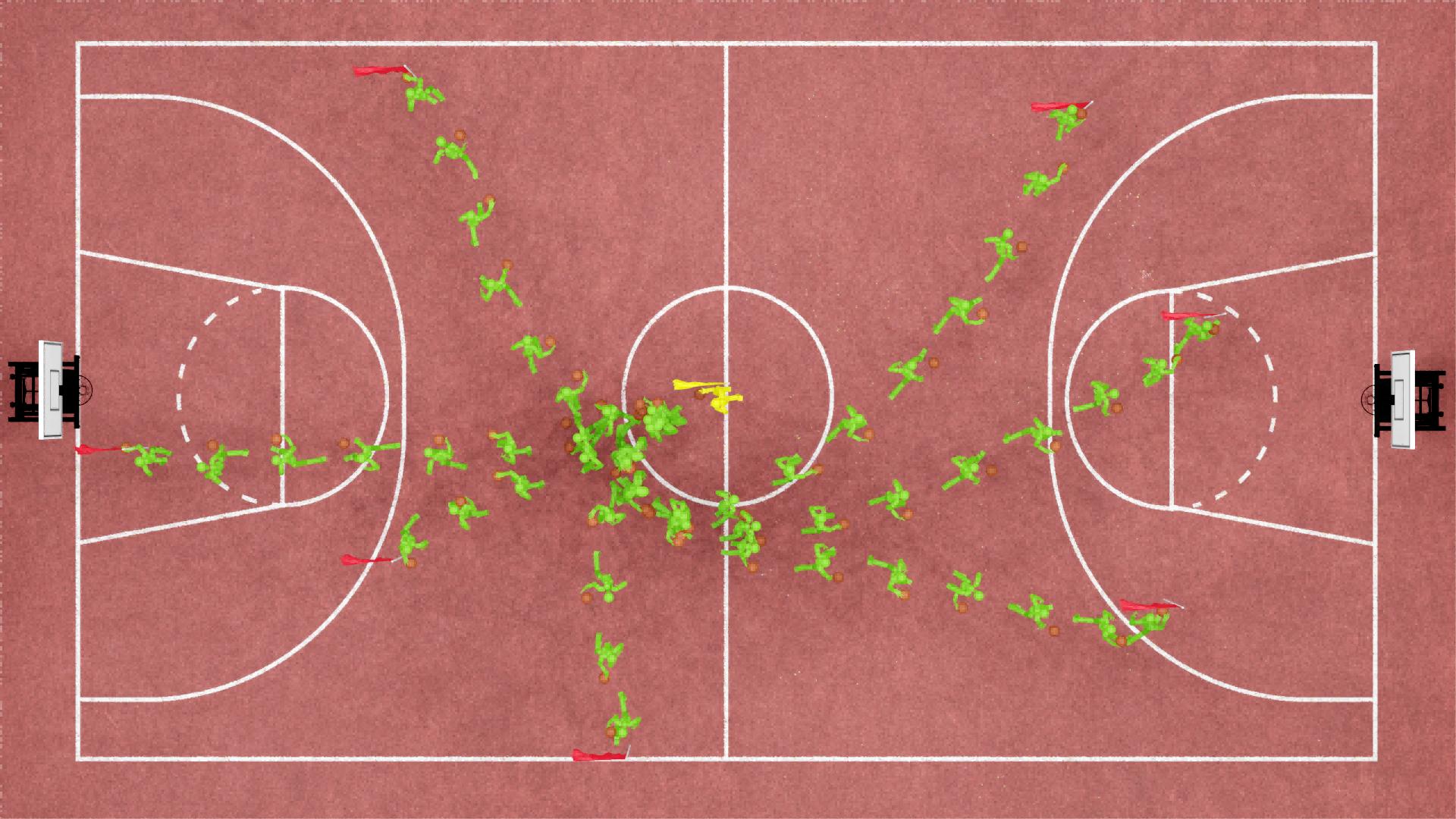}
    \caption{Heading}
  \end{subfigure}
  \begin{subfigure}[b]
  {0.51\columnwidth}
  \includegraphics[width=\linewidth]{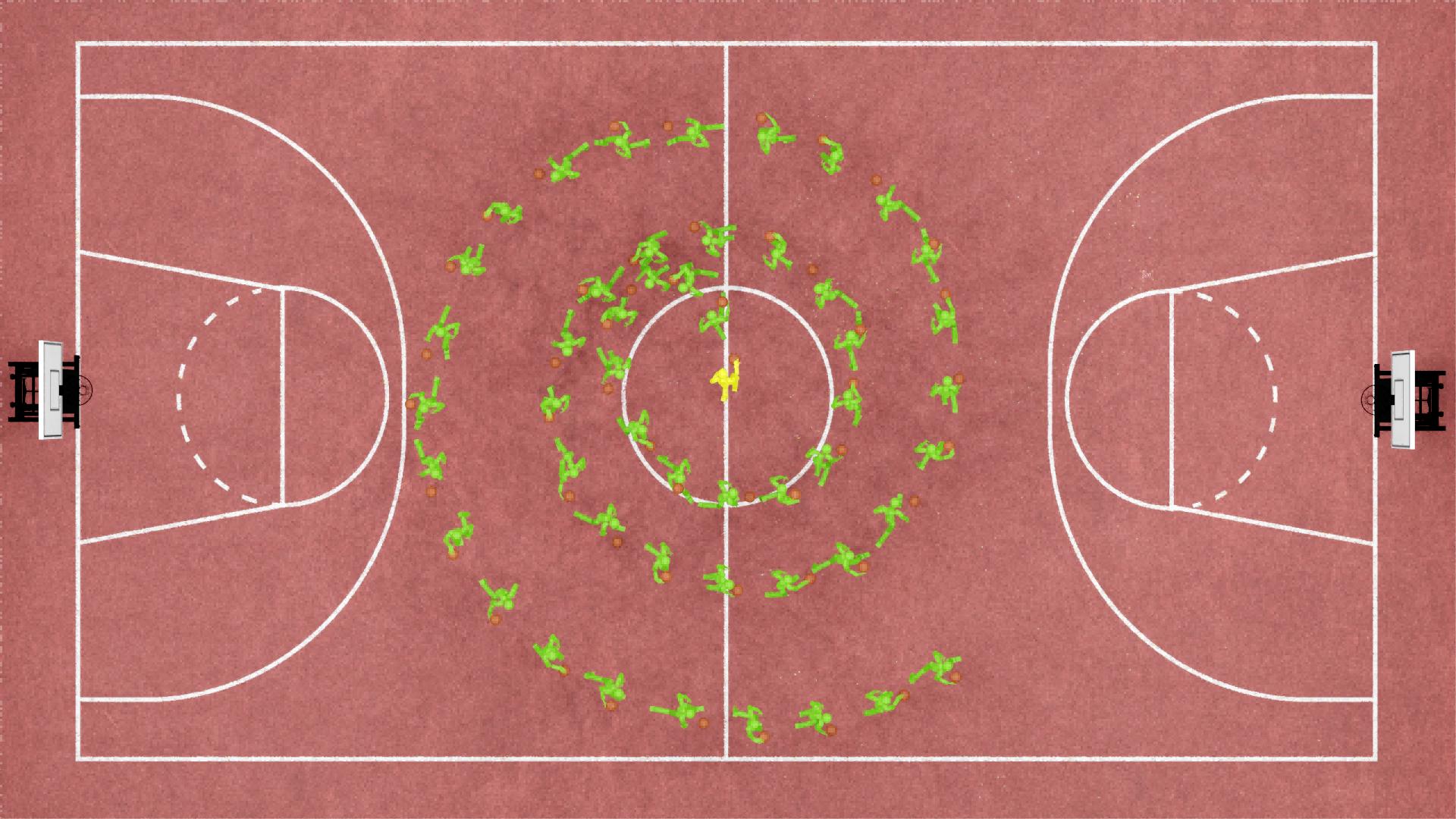}
    \caption{Circling}
  \end{subfigure}
    \caption{Our method supports training a single IS policy under a unified configuration to acquire various interaction skills. These interaction skills can be flexibly switched, as illustrated in (a), where yellow denotes shot, blue denotes pickup, and green denotes turnaround layup. Complex, long-horizon tasks can be easily achieved by training a high-level controller (HLC) to manage switching of the learned interaction skills: (b) scoring from random positions (c) dribbling to target locations, and (d) dribbling along an expanding radius.
    }
    \vspace{-0.2cm}
    \label{fig: hlc}
\end{figure*}

\section{Experiments}

We conduct extensive experiments to evaluate the performance of SkillMimic in learning interaction skills and its application for high-level tasks. Our evaluation consists of two main parts: (1) assessing the capability of SkillMimic in learning interaction skills, including comparisons with existing methods and ablation studies, and (2) demonstrating the effectiveness of reusing acquired interaction skills to learn high-level tasks. Details of the experimental settings and additional experiments on skill switching and robustness are provided in the appendix.

\subsection{Metrics}
\label{sec: metrics}
We consider the following quantitative metrics:
\begin{itemize}
\item \textbf{Accuracy.}
The overall accuracy of HOI imitation, abbreviated as \textit{Acc}., is defined per frame, and deems imitation accurate when the object position and body position errors are both under the thresholds and the contacts are correct. The object threshold is defined as 0.2$m$. The body threshold is defined as 0.1$m$. The \textit{Acc}. is calculated by averaging the values of all frames. 

\item \textbf{Position Error.}
The Mean Per-Joint Position Error (MPJPE) of the body ($E_\text{b-mpjpe}$) and object ($E_\text{o-mpjpe}$) is used to evaluate the positional imitation performance (in $mm$) following ~\cite{luo2023perpetual}. 

\item \textbf{Contact Error.}
The contact error $E_\text{cg}$, ranging from 0 to 1, is defined as $\frac{1}{N}\sum_{t=1}^{N}\text{MSE}(\boldsymbol{s}_{t}^{cg},\hat{\boldsymbol{s}}_{t}^{cg})$,
where N is the total frames of the reference HOI data. 

\item \textbf{Success Rate.}
The execution success rate of an interaction skill. Detailed definitions of success criteria for different skills are provided in the appendix.
\end{itemize}

\subsection{Evaluating Interaction Skill Learning}
Since our approach is the first to learn interaction skills from HOI demonstrations, there are no direct benchmarks for comparison. Therefore, we adapt reward strategies commonly used in locomotion imitation, i.e., DeepMimic~\cite{DeepMimic} and AMP~\cite{amp}, to an object-inclusive setting. 
Specifically, we adapt our reward to the styles of DeepMimic~\cite{DeepMimic} and AMP~\cite{amp} while keeping the other components unchanged for a fair comparison. 
We denote these variant versions as DeepMimic* and AMP*.
In Tab.~\ref{fig: LLC succ}, we present the success rate on 4 typical basketball skills from BallPlay-M, where our method achieves state-of-the-art performance in learning basketball skills compared with variants using DeepMimic or AMP style rewards. We guess that AMP-like rewards are ineffective for learning precise interactions due to their coarse-grained supervision signals.

In Tab.~\ref{tab: ablation of HOI imitation}, we do ablation on reward design, presenting detailed metrics of HOI imitation across two diverse datasets, GRAB \cite{GRAB2020} and BallPlay-V. 
The results indicate that the absence of reward multiplication leads to imbalanced sub-reward learning, e.g., favoring body motion ($E_\text{b-mpjpe}$) over interactions. The absence of CGR leads to kinematic local optima. In contrast, our complete setting (SkillMimic) yields significantly higher performance on interaction-related metrics ($\textit{\ Acc.}$, $E_\text{o-mpjpe}$, and $E_\text{cg}$).

\label{ablation:data_scale}
\subsubsection{Ablation on Contact Graph Reward} 
We conduct ablation experiments on diverse datasets to comprehensively evaluate the impact of Contact Graph Reward (CGR) on interaction skill learning.  Fig.~\ref{fig: cgr} illustrates the qualitative results of ablating CGR on the GRAB and BallPlay-V datasets. It can be clearly observed that without CGR, the humanoid often resorts to incorrect contacts to achieve more stable object control, such as using its head to push the ball, clamping the ball with its hands and legs, or using its hands to support itself on a table for balance. These examples illustrate typical kinematic local optima. By incorporating CGR, these local optima are effectively eliminated, resulting in accurate interactions. Additionally, the result in Tab.~\ref{tab: ablation of HOI imitation} also shows that, without CGR, the contact error ($E_\text{cg}$) significantly increases, leading to kinematic local optima and causes a noticeable decline in overall imitation accuracy. These results underscore the critical importance of CGR for learning correct interactions.

\subsubsection{Data Scale and Generalization Performance} 
To examine how the generalization performance grows as the data scale increases, we conduct ablation experiments on the pickup skill.
We create four datasets by randomly selecting pickup clips from BallPlay-M, containing 1, 10, 40, and 131 clips, respectively. We train policy on each dataset with approximately 3.2 billion samples using SkillMimic.

To evaluate the generalization performance, we randomly place balls within a circular area ranging from 1 to 5 meters around the humanoid—this distribution significantly exceeds the coverage of the reference data. Fig.~\ref{fig: skill diversity} and Fig.~\ref{fig: skill generalization} show the results. We can see that as the data scale increases, pickup performance improves significantly due to broader state transition coverage in larger datasets. 
This indicates our method's strong scalability and potential for large-scale learning. It is worth noting that we do not incorporate any additional designs for robustness and generalization.

\subsection{Evaluating High-level Tasks}

We conduct experiments on four high-level tasks described in Sec.~\ref{sec: reuse skills}. We employ a single IS policy as the interaction skill prior, trained using SkillMimic for around 4.5 billion samples across seven skills: pickup, layup, turnaround layup, dribble left, dribble right, dribble forward, and shot. During training, the IS policy is fixed, and only the HLC is trained. 
For tasks such as throwing, heading, and circling, we train for around 0.4 billion samples. For the more challenging scoring task, we train for around 1.2 billion samples.
To objectively evaluate the performance of our method, we compared it against three baseline methods: (1) learning from scratch using PPO \cite{schulman2017proximal}, (2) learning with body motion priors using ASE \cite{ase}, and (3) learning with body and object motion priors using ASE (denoted as ASE*). The low-level controllers of ASE and ASE* are trained using the same training steps as our IS policy. For task training, all methods are trained using identical task rewards and simulation steps for fair comparisons.

\begin{table}[t]
\centering
\resizebox{0.95\linewidth}{!}{%
\begin{tabular}{l|cccc}
\toprule
\multirow{2}{*}{Method} &  \multicolumn{4}{c}{Success Rate$\uparrow$} \\ 
\cmidrule(l){2-5}
& Heading & Circling & Throwing & Scoring \\
\midrule
PPO\cite{schulman2017proximal} &  0.70\% & 11.14\% & 0.00\% & 0.00\% \\
ASE\cite{ase} & 0.19\% & 4.37\% & 0.00\% & 0.00\% \\
ASE*\cite{ase} & 0.31\% & 7.21\% & 0.00\% & 0.00\%  \\
SkillMimic (ours) &  \textbf{93.04\%} & \textbf{79.92\%} & \textbf{93.40\%}& \textbf{80.25\%}\\
\midrule
\end{tabular}}
    \caption{Success rates on 4 high-level basketball tasks. Both PPO (learn from scratch), ASE (using locomotion prior), and ASE* (using body and object motion prior) fail to converge on all tasks. In contrast, leveraging interaction skill priors acquired through SkillMimic, our method effectively learns these difficult tasks.
    }
\label{tab: hlc}
\end{table}

Fig.~\ref{fig: hlc} (a) showcases an example of manually controlling the pre-trained IS policy to perform skill switching. Fig.~\ref{fig: hlc} (b) shows several trajectories of the HLC performing skill controls to accurately deliver the ball to the target basket. Fig.~\ref{fig: hlc} (c) \& (d) present a top-down view of the HLC controlling the humanoid to dribble the ball toward target locations and along an expanding radius. 
As shown in Tab.~\ref{tab: hlc}, PPO and ASE yield low success rates because the tasks are too challenging to master without interaction priors. The failure of ASE* can be traced back to its low-level controller's inadequate ability to imitate interaction skills. This is primarily because adversarial rewards fail to provide effective guidance for precise interaction learning, which is similar to the failure mode observed in AMP*. In contrast, by utilizing the interaction skills learned through SkillMimic, we greatly improve the performance of learning complex high-level interaction tasks.

\section{Discussion and Future Work}

We presented a novel framework for learning diverse interaction skills with a unified HOI imitation reward using reinforcement learning. Our approach demonstrates multi-skill mastery, skill reusability, and clear data scaling benefits, highlighting the significant potential of HOI imitation for learning complex interaction skills. However, the full application of this paradigm requires further exploration. Several key areas deserve particular attention:

Firstly, exploring more general HOI skill learning within a single policy, such as household tasks or other types of sports, presents challenges in handling generalization across different object geometries. Moreover, perceiving multiple objects simultaneously requires a more general environmental perception approach, rather than directly obtaining privileged information for each object.

Secondly, incorporating finer-grained control conditions, such as tracking \cite{luo2023perpetual}, could enable more flexible control and generalization capabilities as HOI data scales up.

Thirdly, the difficulty of collecting HOI data highlights the importance of exploring methods to learn robust interaction skills with limited HOI data, as well as investigating HOI data augmentation and generation techniques.

Lastly, applying this technology to real robots would be highly valuable, such as enabling real-world humanoid robots to play basketball games. This endeavor could involve various challenges, including HOI data retargeting and sim-to-real transfer issues.

\section*{Acknowledgment}
We would like to thank Mingyang Su for his valuable assistance in motion capture data collection and Shunlin Lu for helpful discussions. This work was supported in part by Unitree Robotics, International Digital Economy Academy (IDEA), and the Innovation and Technology Fund of HKSAR (GHX/054/21GD).
{
    \small
    \bibliographystyle{ieeenat_fullname}
    \bibliography{main}
}
\clearpage
\setcounter{page}{1}
\maketitlesupplementary

\section{The BallPlay Dataset}
\label{app: ballplay}

To address the scarcity of basketball HOI data and facilitate research on basketball skill learning, we introduce two datasets: one based on monocular vision estimation and the other using multi-view optical motion capture systems.

\subsection{BallPlay-V}
\begin{figure}[t]
  \centering
  \includegraphics[width=1\linewidth]{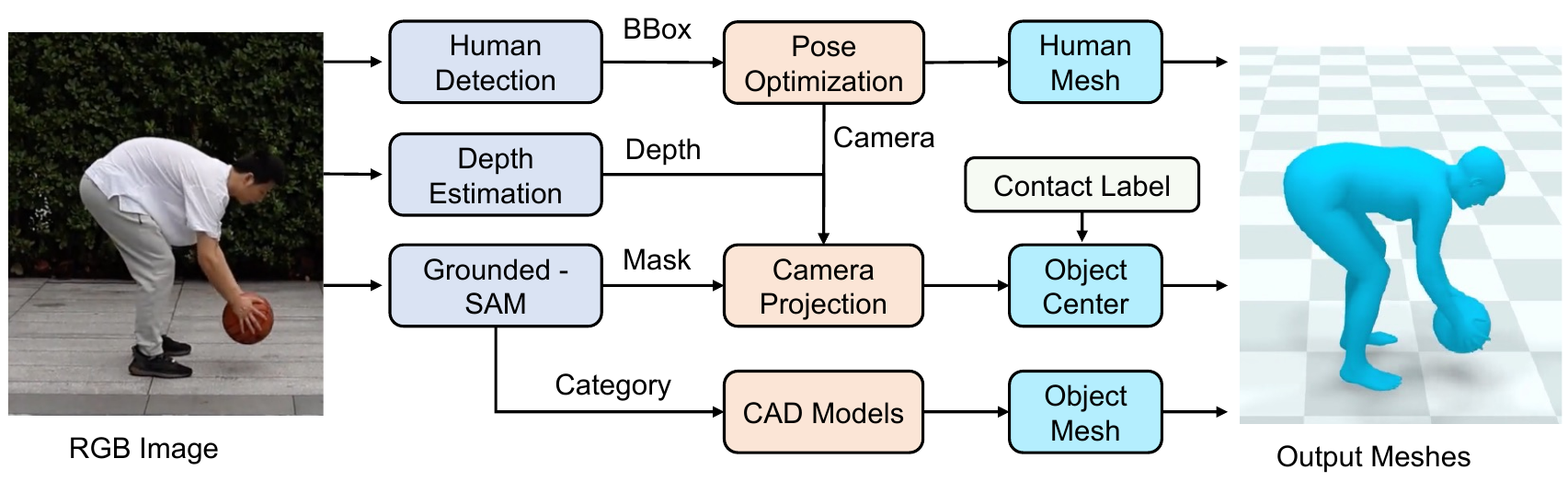}
  \caption{Annotation pipeline of BallPlay-V. Given an image, the pipeline estimates the human SMPL-X \cite{SMPL-X} parameters and object translations. The object is predefined as CAD models.}
\label{fig:annotation} 
\end{figure}

\begin{figure}[h]
  \centering
  \includegraphics[width=1\linewidth]{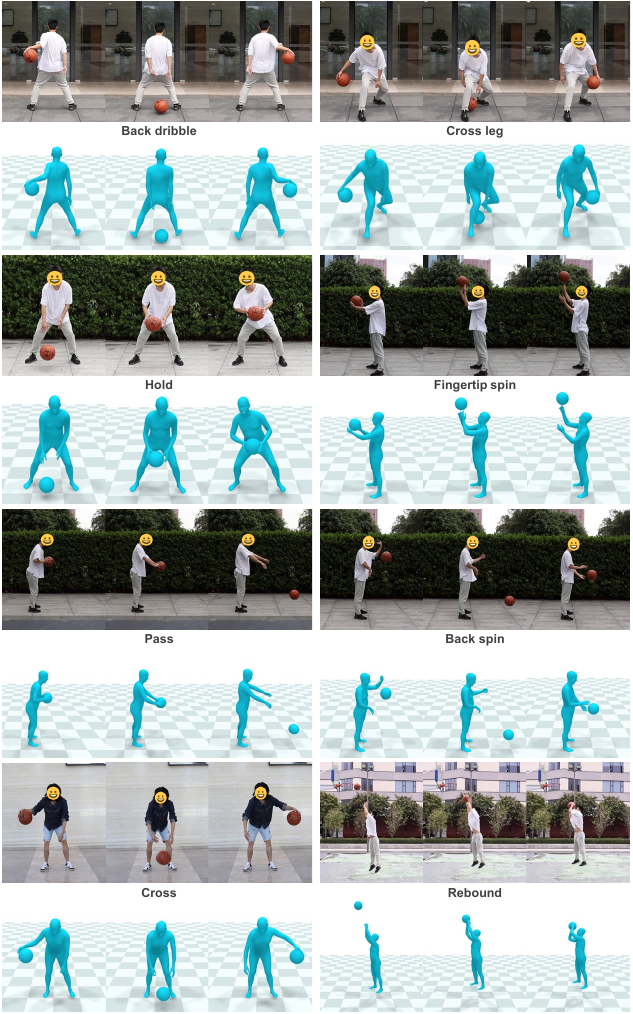}
  \caption{The BallPlay-V dataset. We show the eight HOI demonstrations of high-dynamic basketball skills. For each skill, the \textbf{upper} rows show the real-life videos, and the \textbf{lower} rows give the estimated whole-body SMPL-X human model and object mesh.}
\label{fig: ballplay} 
\end{figure}

The BallPlay-V dataset applies a monocular annotation solution to estimate the high-quality human SMPL-X \cite{SMPL-X} parameters and object translations from RGB videos. However, annotating these videos with high-speed and dynamic movements and complex interactions in the 3D camera coordinate is quite challenging. 
Inspired by the whole-body annotation pipeline of Motion-X~\cite{lin2023motionx}, our automatic annotation additionally introduces depth estimation~\cite{bhat2023zoedepth}, semantic segmentation~\cite{ren2024grounded}, to obtain high-quality whole-body human motions and ball motions, as illustrated in Fig.~\ref{fig:annotation}. The BallPlay-V dataset contains eight basketball skills, including \emph{back dribble, cross leg, hold, fingertip spin, pass, backspin, cross}, and \emph{rebound}, as shown in Fig.~\ref{fig: ballplay}.

\subsection{BallPlay-M}

Although BallPlay-V can acquire HOI data conveniently from RGB videos, its accuracy is limited due to errors of monocular depth estimation. Additionally, it struggles with occlusion issues, making it difficult to capture complex layup and dribbling data. To achieve more comprehensive and accurate basketball data, we create the BallPlay-M dataset using a optical motion capture system. During the capture process, optical markers are attached to both the player and the basketball to track body and ball movements. The player also wears gloves equipped with inertia sensors to estimate finger movements. Consequently, the player is parameterized as a skeleton with 52 joints (156 DOFs). We calculate and record the root rotation, root translation, joint positions, and joint rotations sequentially. The ball is parameterized as a sphere, with its rotation and center position recorded. All data are captured at 120 fps.

We collect a total of 251,656 frames of raw data at a frame rate of 120 fps, amounting to approximately 35 minutes of diverse basketball interactions. From these raw data, we extract and annotate a subset for learning basketball skills. To aid reader comprehension, we provide a coarse categorization of the labeled subset in Tab.~\ref{tab: skilldist}. It is important to note that this coarse classification is intended for illustrative purposes only. In reality, we have annotated the skills with greater specificity. For example, the \textit{dribble} category encompasses various distinct skills, such as \textit{dribble forward}, \textit{dribble right}, \textit{back dribble}, and so on.

\begin{table}[t]
\centering  
\resizebox{0.9\linewidth}{!}{
\begin{tabular}{|l|c|c|c|}
\hline
{\bf Coarse-Grained Classification} & {\bf Clips} & {\bf Frames} & {\bf FPS} \\ \hline
    Pick Up &  $105$&  $21,232$ & $60$ \\ \hline
    Catch &  $12$&  $1,341$ & $60$ \\ \hline
    Rebound &  $31$&  $2,175$ & $60$ \\ \hline
    Dribble &  $59$&  $7,747$ & $60$ \\ \hline
    Shot &  $23$ &  $3,212$ & $60$ \\ \hline
    Pass &  $10$&  $604$ & $60$ \\ \hline
    Layup &  $19$ &  $2,765$ & $60$ \\ \hline
    Getup & $16$&  $3,170$ & $60$ \\ \hline
    Misc. &  $24$ &  $4,812$ & $60$ \\ \hline
    \textbf{Raw Data} &  - &  $251,656$ & $120$ \\ \hline
\end{tabular}
}
\caption{The composition of BallPlay-M dataset. We extracted and annotated a subset from approximately 35 minutes of raw data for skill learning purposes. We show the coarse categorization of the annotated subset in the above table.
}
\label{tab: skilldist}
\end{table}

\section{Additional Results and Experiments}

\subsection{Additional Qualitative Results}
\begin{figure*}[t]
  \centering
  \vspace{-0.2cm}
  \begin{subfigure}[b]{1\columnwidth} \includegraphics[width=\linewidth]{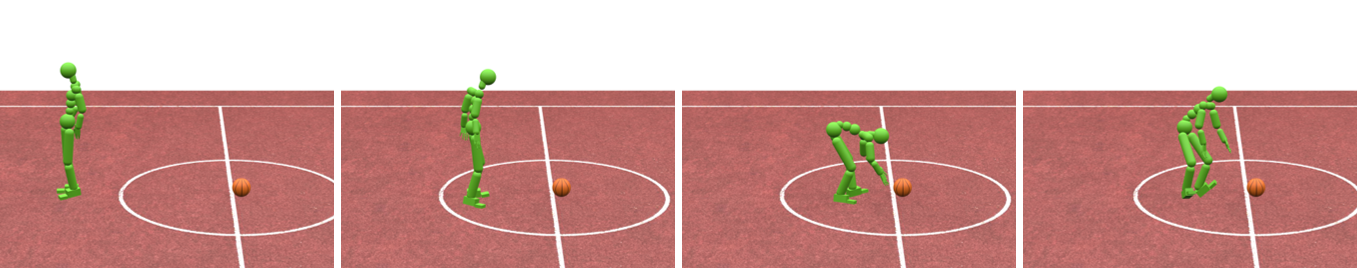}
    \caption{DeepMimic* on Pick Up}
  \end{subfigure}
  \begin{subfigure}[b]{1\columnwidth}
  \includegraphics[width=\linewidth]{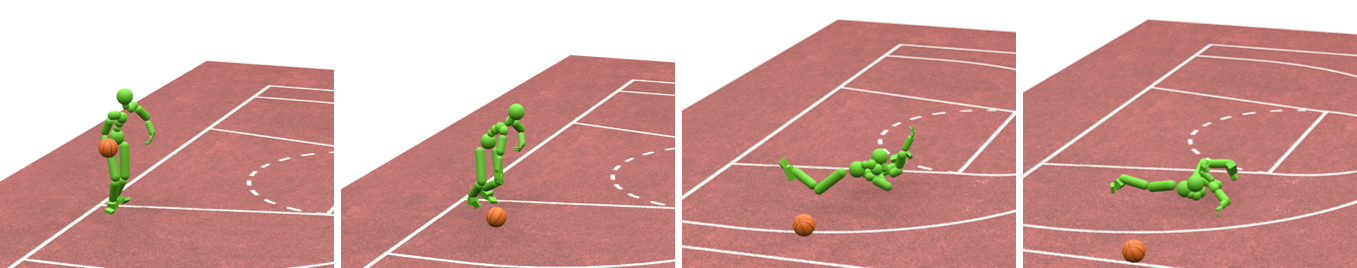}
    \caption{DeepMimic* on Dribble Forward}
  \end{subfigure}
  \\
  \begin{subfigure}[b]
  {1\columnwidth}
  \includegraphics[width=\linewidth]{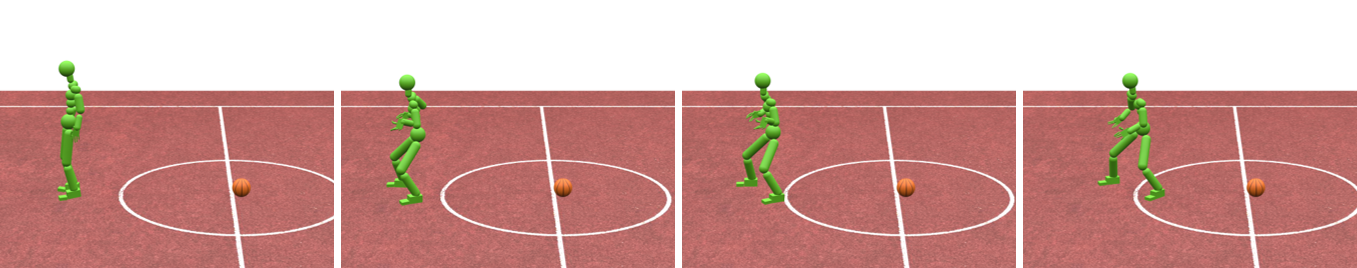}
    \caption{AMP* on Pick Up}
  \end{subfigure}
  \begin{subfigure}[b]
  {1\columnwidth}
  \includegraphics[width=\linewidth]{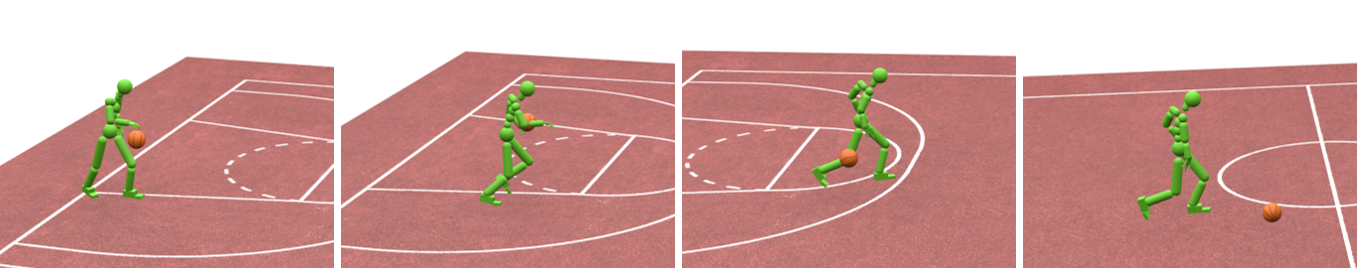}
    \caption{AMP* on Dribble Forward}
  \end{subfigure}
  \\
  \begin{subfigure}[b]
  {1\columnwidth}
  \includegraphics[width=\linewidth]{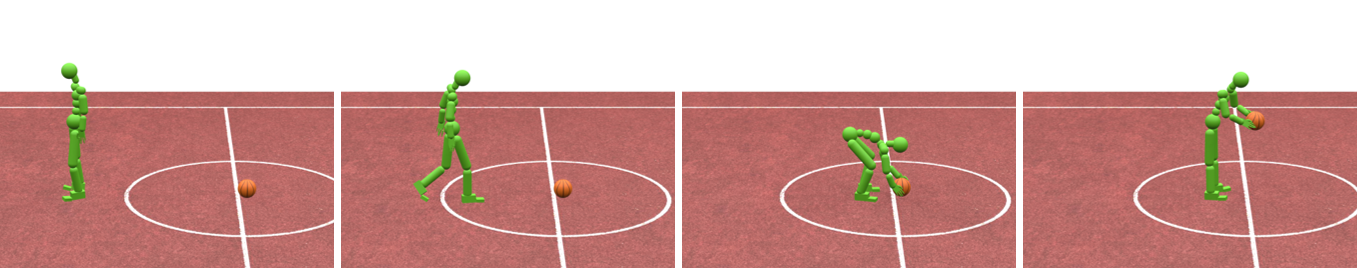}
    \caption{SkillMimic on Pick Up}
  \end{subfigure}
  \begin{subfigure}[b]
  {1\columnwidth}
  \includegraphics[width=\linewidth]{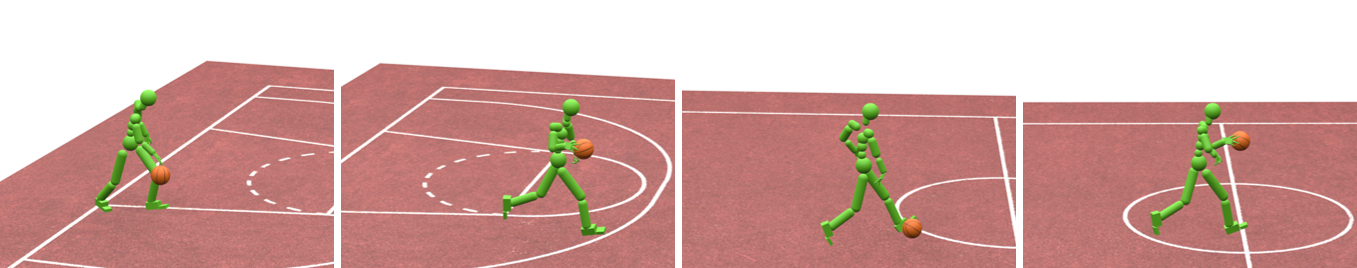}
    \caption{SkillMimic on Dribble Forward}
  \end{subfigure}
   \caption{
    Comparisons on imitation learning of Interaction Skills. 
    Both DeepMimic* and AMP* demonstrate difficulties in simultaneously managing object and body motion learning. For instance, while AMP* can roughly replicate limb movements during dribbling, it struggles to precisely control ball movement. Additionally, AMP* suffers from mode collapse, manifesting as prolonged hesitation during ball pickup attempts. In contrast, our unified HOI imitation reward successfully addresses these limitations, demonstrating superior performance in HOI imitation and establishing the first solution for purely data-driven interaction skill learning.
   }
   \label{fig: comp1}
\end{figure*}

\begin{figure*}[t]
  \centering
  \vspace{-0.2cm}
  \begin{subfigure}[b]{1.0\columnwidth} \includegraphics[width=1\linewidth]{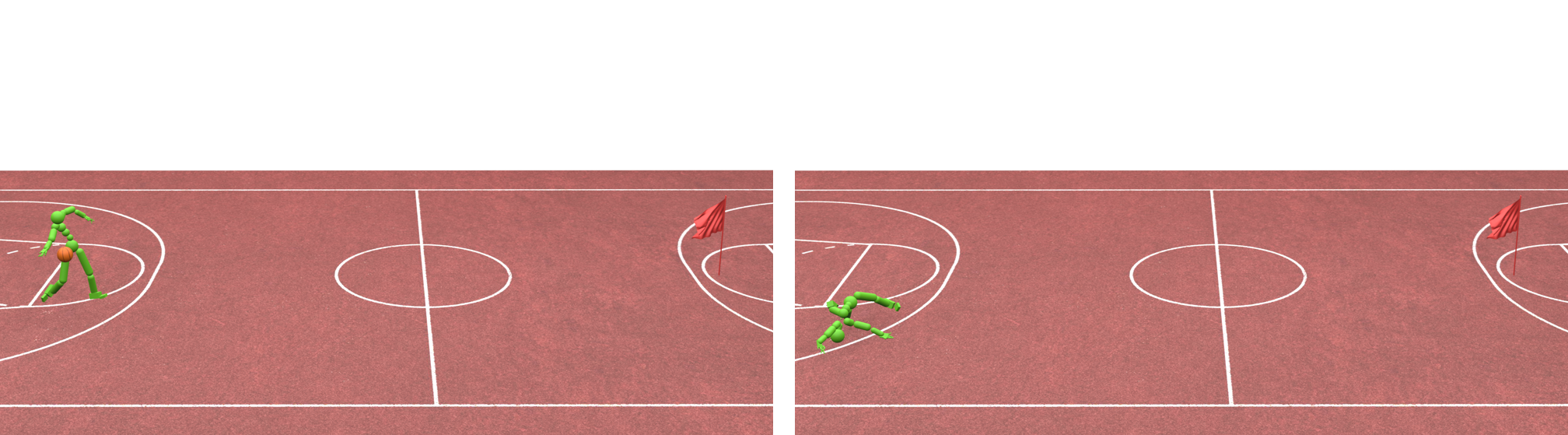}
    \caption{PPO on Heading}
  \end{subfigure}
  \,
  \begin{subfigure}[b]
  {1.0\columnwidth}
  \includegraphics[width=1\linewidth]{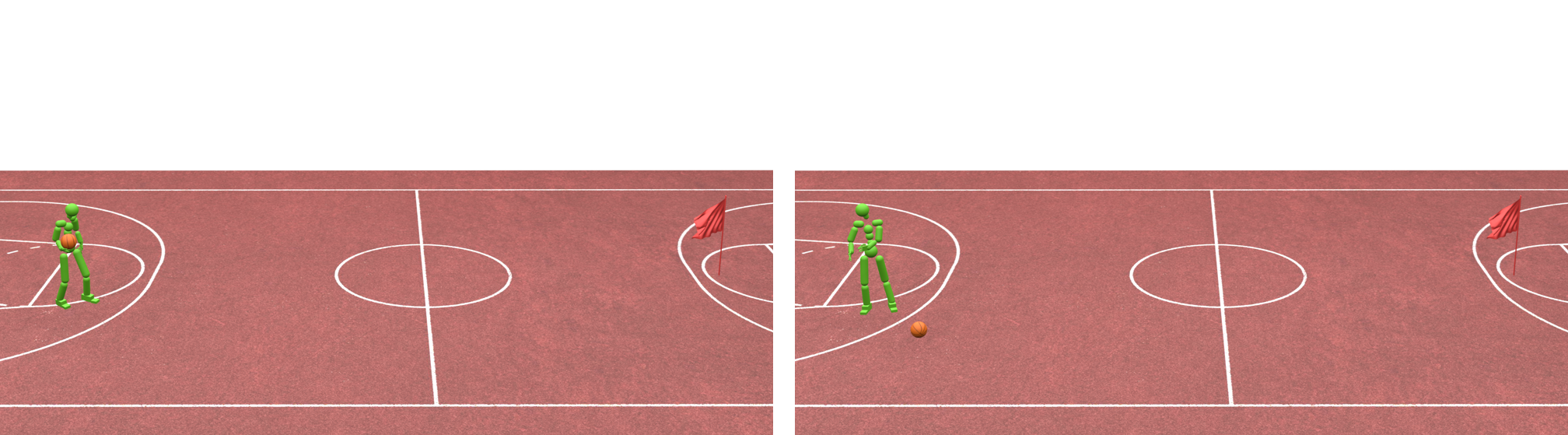}
    \caption{ASE on Heading}
  \end{subfigure}
  \\
  \vspace{-0.1cm}
  \begin{subfigure}[b]
  {1.0\columnwidth}
  \includegraphics[width=1\linewidth]{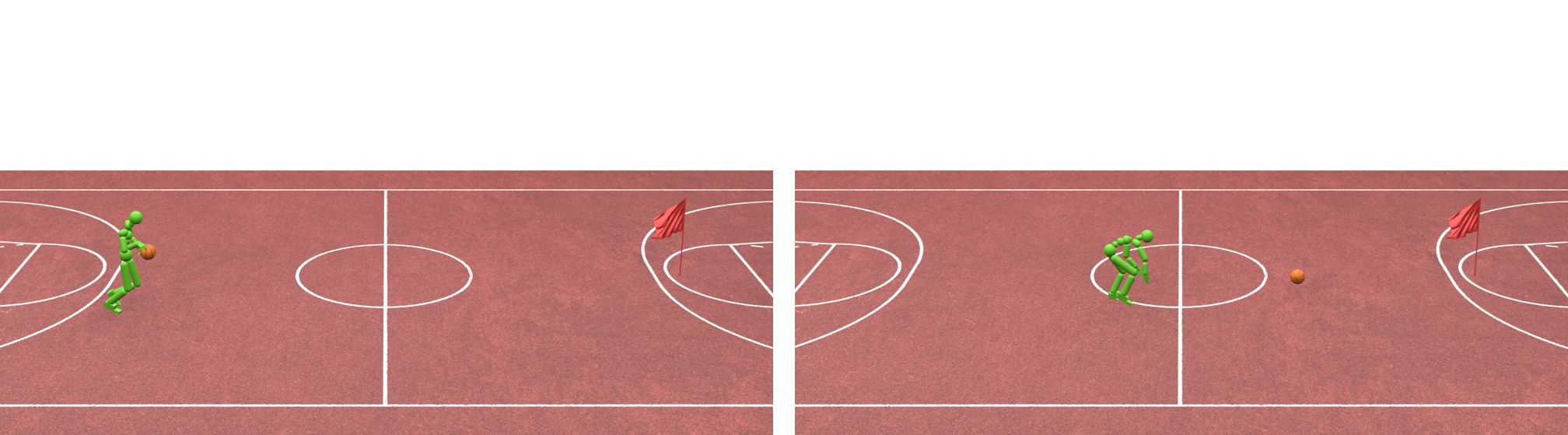}
    \caption{ASE* on Heading}
  \end{subfigure}
  \,
  \begin{subfigure}[b]
  {1.0\columnwidth}
  \includegraphics[width=1\linewidth]{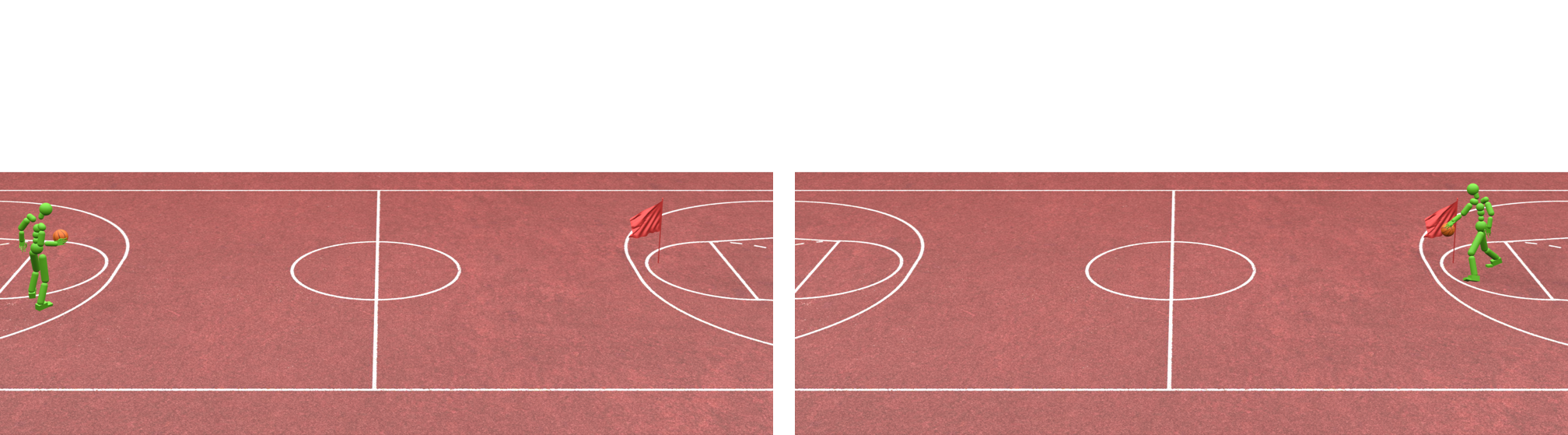}
    \caption{Ours on Heading}
  \end{subfigure}
  \\
  \vspace{-0.1cm}
  \begin{subfigure}[b]{1.0\columnwidth}
  \includegraphics[width=1\linewidth]{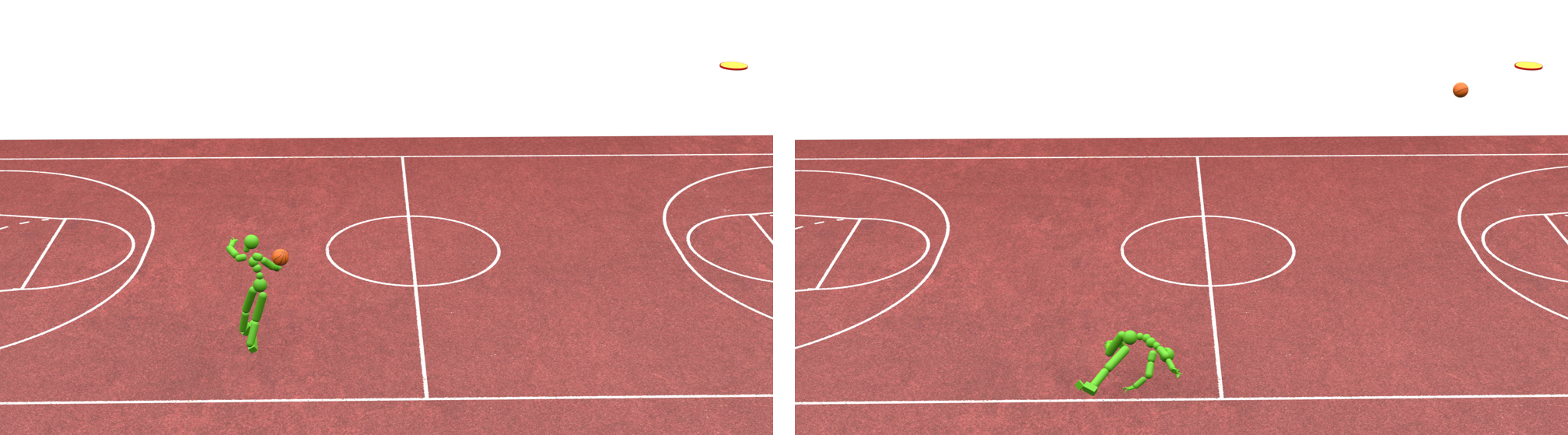}
    \caption{PPO on Scoring}
  \end{subfigure}
  \,
  \begin{subfigure}[b]
  {1.0\columnwidth}
  \includegraphics[width=1\linewidth]{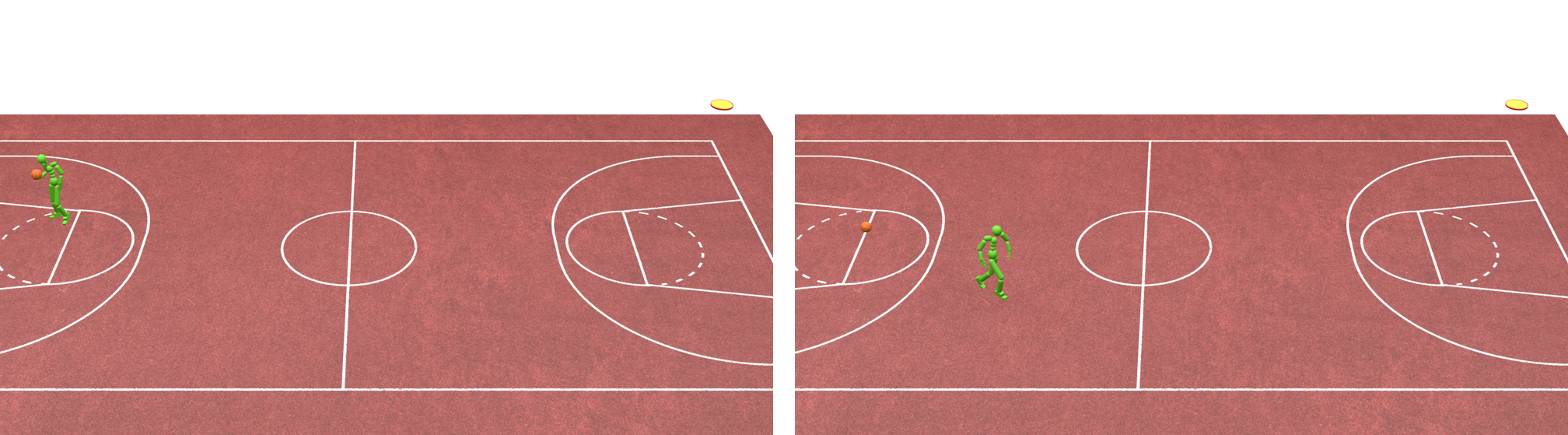}
    \caption{ASE on Scoring}
  \end{subfigure}
  \\
  \vspace{-0.1cm}
  \begin{subfigure}[b]
  {1.0\columnwidth}
  \includegraphics[width=1\linewidth]{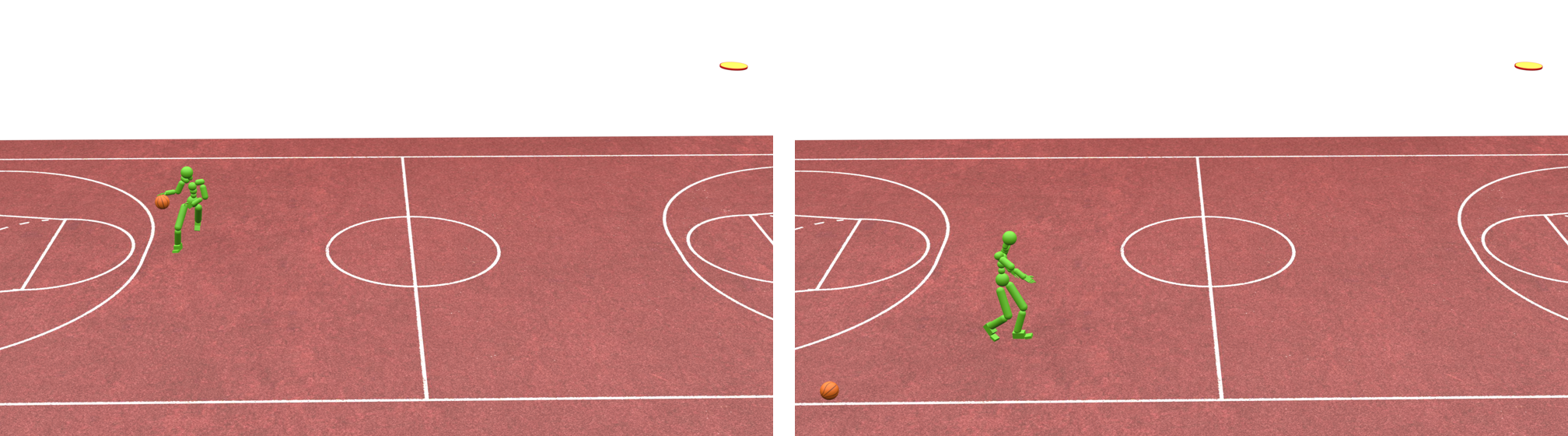}
    \caption{ASE* on Scoring}
  \end{subfigure}
  \,
  \begin{subfigure}[b]
  {1.0\columnwidth}
  \includegraphics[width=1\linewidth]{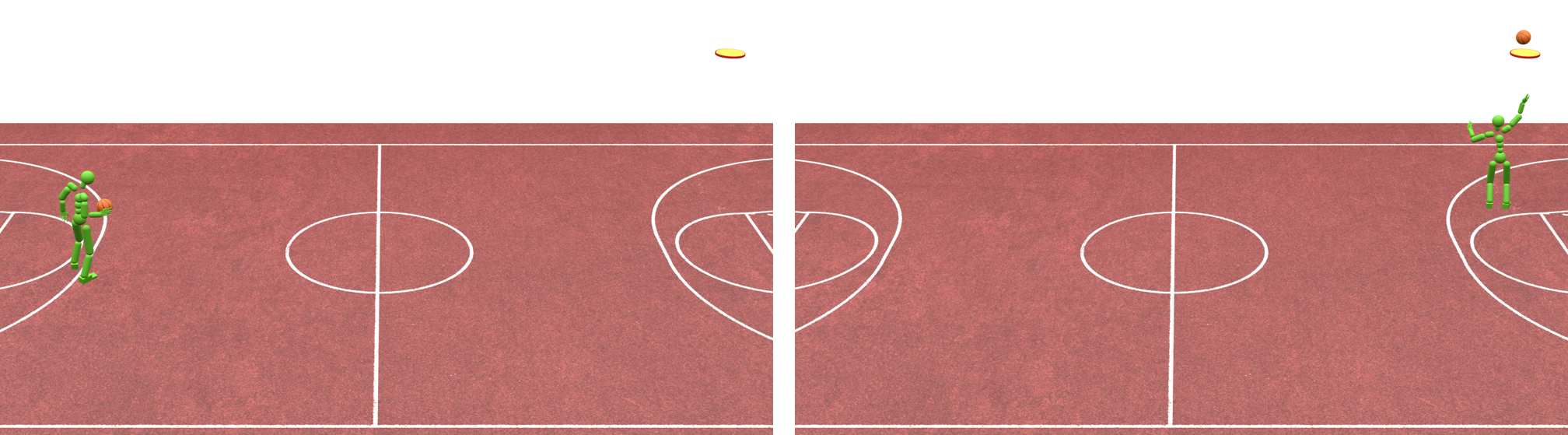}
    \caption{Ours on Scoring}
  \end{subfigure}
   \caption{Comparisons on high-level basketball tasks. Left: start status. Right: end status. The high-level task rewards are extremely sparse as they only depend on object states, making convergence challenging for training from scratch (denoted as PPO). Even with locomotion skill priors, achieving convergence remains difficult (denoted as ASE). Instead, our approach first trains an IS policy through SkillMimic to acquire basic basketball interaction skills, then learns a HLC to effectively compose these interaction skills for high-level tasks. We also adapt ASE to enable direct imitation of human-object interactions in its LLC (denoted as ASE*). However, the coarse-grained nature of GAIL rewards prevents the policy from learning precise interactions, consequently hindering HLC convergence. 
   }
   \label{fig: comp3}
   \vspace{-0.2cm}
\end{figure*}

\begin{figure}[t]
  \centering  \includegraphics[width=\linewidth]{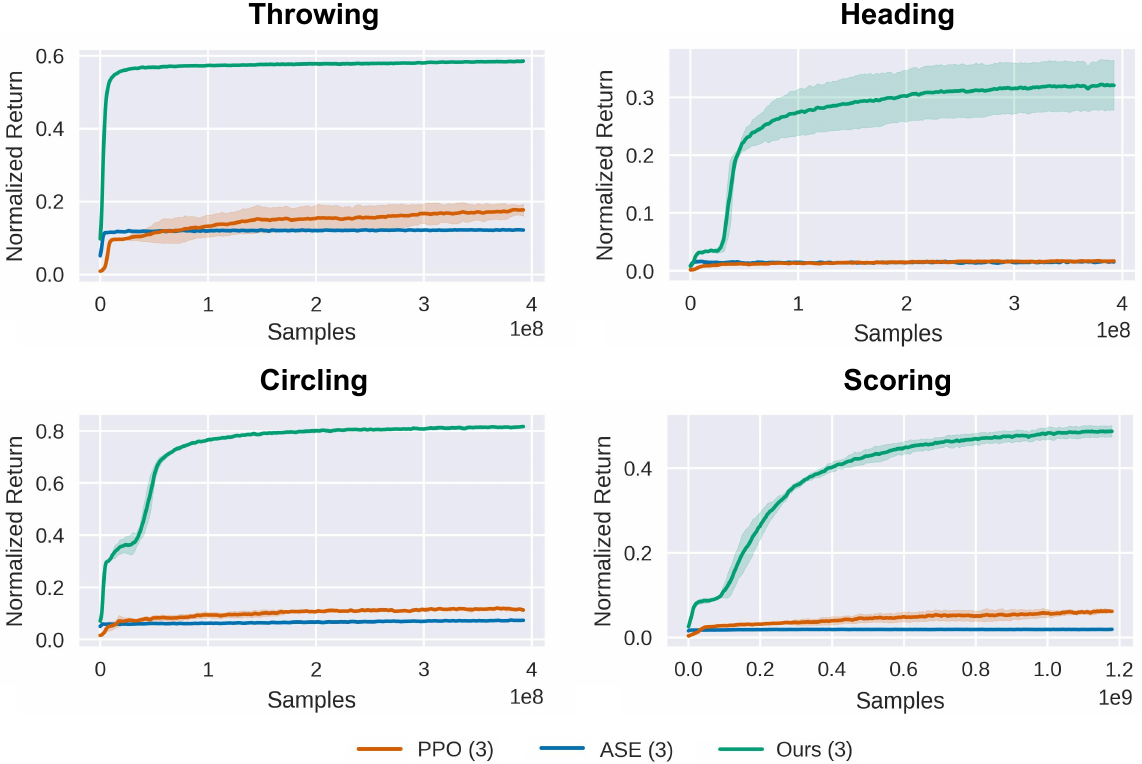}
  \caption{Learning curves of different methods on 4 high-level basketball tasks. On these challenging tasks, sparse task rewards struggle to converge when training from scratch (PPO) or using locomotion priors (ASE), whereas our method achieves rapid convergence by using interaction skill prior.} 
\label{fig: HLC reward}
\end{figure}

We present comprehensive video results on our project page, demonstrating various aspects including basketball skill acquisition, high-level task execution, skill transitions, and comparative methods. Here we offer an in-depth comparison with variant methods, where detailed illustrations are provided in Fig.~\ref{fig: comp1}, Fig.~\ref{fig: comp3}, and Fig.~\ref{fig: HLC reward}. 

\subsection{Skill Robustness Against Physical Properties}
To evaluate the robustness of the skills learned through SkillMimic, we conduct three perturbation tests on the dribble forward and pickup skills during inference: (1) varying the ball radius from 0.5$\times$ to 1.5$\times$ the default size; (2) altering the ball density from 0.1$\times$ to 6$\times$ the default; and (3) changing the ball restitution from 0.5$\times$ to 1.5$\times$ the default. The success rate was averaged across 1000 parallel environments. The quantitative results, presented in Tab.~\ref{tab: robust}, demonstrate our method's robustness against variations in physical properties and external disturbances.

\subsection{Ablation on Mixed Skill Training.}
We conduct the following experiment to analyze the effect between different skills when learned together. From BallPlay-M, we select one clip for the layup skill, one clip for the dribble left skill, one clip for the dribble right skill, and two clips for the dribble forward skill. These skills are commonly used and combined in real-world basketball games.
We first train four individual policies for each skill independently, with each policy trained for around 0.65 billion samples. As a control, we then train these four skills using a single policy, and report the results when trained for around 0.65 billion samples (the same as the individual training) and 2.6 billion samples (4 times of the individual training, but the average sampling number for each skill is the same as in the individual training). During testing, we compare the success rates of executing each skill independently and transitioning between skills. Specifically, for testing each skill, we use the skill clips for Reference State Initialization (RSI) and execute the corresponding skill. For testing skill switching, we use the source skill clips for RSI and execute the target skill. 

Success rates are calculated as described in Sec.~\ref{sec: success rate}. Tab.~\ref{tab: mixtrain} presents the quantitative results. Despite equal sampling for each skill in both sets of experiments, mixed training shows a significant improvement in the success rates of individual skills and an even more substantial improvement in skill switching. It should be noted that while reward convergence is slightly faster in individual training, this approach is susceptible to overfitting. For example, the DL skill demonstrates excellent convergence of its reward, but due to the inadequate state cycle in the reference data, the DL skill trained independently tends to fall after a few dribbling steps, resulting in a zero success rate in sustained operations. Conversely, mixed training allows for cross-learning from other skills, thereby significantly enhancing the success rate of the dribble left skill. A similar phenomenon is observed in skill switching, where the reference data lacks examples of skill switches. Mixed training enables the policy to adapt to the state distributions of all skills, facilitating zero-shot skill switching during tests. These findings not only demonstrate that SkillMimic can support a single policy to learn diverse skills but also underscore the importance of mixed training in enhancing skill generalization and robustness.

\begin{table*}[t!]
\centering
\resizebox{\linewidth}{!}{%
\begin{tabular}{c|cccccc|cccccc|cccccc}
\toprule
\multicolumn{1}{c}{} & \multicolumn{6}{c}{\text{Ball Radius} } & \multicolumn{6}{c}{\text{Ball Density}}& \multicolumn{6}{c}{\text{Ball Restitution}}
\\ 
\midrule
Skill  &   $0.5\times$ &   $0.7\times$ &   $0.9\times$ &    $1.1\times$   &   $1.3\times$ &   $1.5\times$   &   $0.1\times$ &   $0.4\times$ &    $0.7\times$ &    $2\times$   &   $3\times$ &   $4\times$     &   $0.6\times$ &   $0.8\times$ &    $1.2\times$ &    $1.4\times$   &   $1.6\times$ &   $1.8\times$  \\ \midrule
Dribble Forward   &  {0.0$\%$} & {29.0$\%$} & {84.2$\%$}  & {85.5$\%$} & {57.2$\%$} & {0.0$\%$} & {0.1$\%$} & {60.1$\%$}  & {79.5$\%$} & {92.0$\%$} & {33.3$\%$} & {0.0$\%$} & {7.0$\%$}&{87.6$\%$} & {87.0$\%$}& {85.8$\%$}& {76.1$\%$} & {3.64$\%$}\\
Pickup  &  {2.2$\%$} & {58.7$\%$} & {78.7$\%$}  & {79.7$\%$} & {64.1$\%$} & {0.4$\%$} & {12.2$\%$} & {78.3$\%$}  & {79.1$\%$} & {79.1$\%$} & {75.3$\%$} & {17.4$\%$} & {79.0$\%$}&{79.6$\%$} & {78.6$\%$}& {79.6$\%$}& {79.2$\%$} & {78.9$\%$} \\
\midrule
\end{tabular}
}
\vspace{-0.1cm}
    \caption{Impact of varying physical properties on success rates. Models trained with fixed physical attributes were tested by scaling these attributes by a factor. The results show that the interaction skills learned by SkillMimic are robust against minor physical property changes.}
\label{tab: robust}
\end{table*}

\begin{table*}[t!]
\centering
\resizebox{\linewidth}{!}{%
\begin{tabular}{c|cccc|cccc}
\toprule
\multicolumn{1}{c}{} & \multicolumn{4}{c}{\text{Success Rate} \text{on Individual Skills} } & \multicolumn{4}{c}{\text{Success Rate} \text{on Skill Switching}}
\\ 
\midrule
Training  &   \text{Dribble Forward} &   \text{Dribble Left} &   \text{Dribble Right} &    \text{Layup}   &   \text{Dribble Forward}-\text{Left}   &   \text{Dribble Forward}-\text{Right} &   \text{Dribble Forward}-\text{Layup} &    \text{Dribble Left}-\text{Forward}
\\ \midrule
Ind.-1$\times$  &  {41.3$\%$} &  {0.0$\%$} &  {81.0$\%$} &  {95.5$\%$} &  {0.0$\%$} &  {5.14$\%$} &  {8.2$\%$} &  {0.09$\%$}\\
Mixed-1$\times$   &  62.8$\%$ &  4.1$\%$ &  48.14$\%$ &  \textbf{100.0$\%$} &  1.7$\%$ & 8.8$\%$ &  40.5$\%$ &  13.5$\%$\\
Mixed-4$\times$   &  \textbf{87.3$\%$}  &  \textbf{67.9$\%$} &  \textbf{92.6$\%$} &  99.9$\%$ &  \textbf{60.5$\%$} &  \textbf{14.5$\%$} &  \textbf{40.6$\%$} &  \textbf{46.3$\%$}\\
\midrule
\end{tabular}
}
\vspace{-0.1cm}
    \caption{Success rates of skills trained independently versus jointly. Ind. denotes individual training. 1$\times$ denotes 0.65 billion training samples while 4$\times$ denotes 4 times of that. Mixed training significantly improves both individual skill execution and skill switching, demonstrating the effectiveness of SkillMimic in handling diverse interaction skills at once.}
\label{tab: mixtrain}
\end{table*}

\section{Implementation Details}
\subsection{Simulation Settings.}
We use Isaac Gym \cite{makoviychuk2021isaac} as the physics simulation platform. All experiments are trained on a single Nvidia RTX 3090 or 4090 GPU, with 2048 parallel environments. For GRAB and BallPlay-V, both the simulation and PD controller operate at 60 Hz, while the skill policy is sampled at 30 Hz. For BallPlay-M, the simulation and PD controller run at 120 Hz, with the skill policy sampled at 60 Hz. We resample the reference HOI clips to match the skill policy frequency, and the high-level policy is sampled at 20 fps. 
All neural networks are implemented using PyTorch and trained using Proximal Policy Optimization \cite{schulman2017proximal}. 
We use the edge set $\mathcal{E}$ of the CG and calculate the CG edge values by judging the contact force of each CG node. The setting of hyperparameters is fixed for all experiments and can be found in Sec.~\ref{app: hyperparameters}.

For GRAB \cite{GRAB2020} and BallPlay-V, the whole-body humanoid follows the SMPL-X~\cite{SMPL-X} kinematic tree and has a total of 52 body parts and 51$\times$3 DOF actuators where 30$\times$3 DOF is for the hands and 21$\times$3 DOF for the rest of the body. For BallPlay-M, the humanoid model consists of 53 body parts and 52$\times$3 DOF actuators, the hands having 30$\times$3 DOF and the rest of the body having 22$\times$3 DOF. The basketball is modeled as a sphere with a radius of 12 cm, which is close to the size of a real-world basketball. The restitution coefficients for the plane and the ball are set to 0.8 and 0.81, respectively, to ensure the ball's bounce closely resembles real-world basketball behavior. The humanoid's mass is set to match that of a real player. We set the ball's density to 1000 kg/m$^3$ to enhance stability and accelerate training convergence, while other physical parameters remain at their default settings. Despite being trained with fixed physical properties,
our method can withstand a wide range of changes in physical properties during inference, such as variations in the ball's density, radius, and restitution, as shown in Tab.~\ref{tab: robust}.

\subsection{Kinematic Imitation Rewards.} \label{app: Kin Reward}
Kinematic imitation rewards form the basis of the HOI imitation. We design these rewards in four distinct parts: the Body Kinematics Reward $r_{t}^{b}$, the Object Kinematics Reward $r_{t}^{o}$, the Relative Motion Reward $r_{t}^{rel}$, and a Velocity Regularization term $r_{t}^{reg}$.

The Body Kinematics Reward $r_{t}^{b}$ encourages the alignment of the body's movements with the reference data:
\begin{equation}
\begin{aligned}
    r_{t}^{b} = r_{t}^{p}*r_{t}^{r}*r_{t}^{pv}*r_{t}^{rv},
    \label{eq: kin}
\end{aligned}
\end{equation}
where $r_{t}^{p}$, $r_{t}^{r}$, $r_{t}^{pv}$, $r_{t}^{rv}$ are the humanoid position reward, rotation reward, position velocity reward, and angular velocity reward. Each sub-reward is calculated by computing the Mean Squared Error (MSE) with the reference data, followed by a negative exponential normalization. For instance, the calculation for $r_{t}^{p}$ is as follows:
\begin{equation}
\begin{aligned}
    r_{t}^{p} = \text{exp}(-\lambda^{p} *e_{t}^{p}), \quad e_{t}^{p} = \text{MSE}(\boldsymbol{s}_{t}^{p},\hat{\boldsymbol{s}}_{t}^{p}),
    \label{eq: rp}
\end{aligned}
\end{equation}
where $\hat{\boldsymbol{s}}_{t}^{p}$, is the reference humanoid body positions, $\boldsymbol{s}_{t}^{p}$ is the simulated humanoid body positions, $\lambda^{p}$ is a hyperparameter that conditions the sensitivity.

The Object Kinematics Reward $r_{t}^{o}$ ensures the object's movements are consistent with the reference:
\begin{equation}
\begin{aligned}
    r_{t}^{o} = r_{t}^{op}*r_{t}^{or}*r_{t}^{opv}*r_{t}^{orv},
    \label{eq: ro}
\end{aligned}
\end{equation}
where $r_{t}^{op}$, $r_{t}^{or}$, $r_{t}^{opv}$, $r_{t}^{orv}$ are the object position reward, rotation reward, position velocity reward, and angular velocity reward, respectively. The calculation of these sub-rewards resembles Eq.~\ref{eq: rp}.

The relative motion is represented as a vector group, obtained by subtracting the object's position from the key body positions. The calculation of Relative Motion Reward $r_{t}^{rel}$ is also similar to Eq.~\ref{eq: rp} and is effective in constraining the relative motion between the object and key body points to be consistent with the reference.

Lastly, a Velocity Regularization term is employed to suppress high-frequency jitter in the humanoid when it is supposed to be stationary:
\begin{equation}
\begin{aligned}
    r_{t}^{reg} = \text{exp}(-\lambda^{reg} *e_{t}^{acc}), 
    \label{eq: ref}
\end{aligned}
\end{equation}
\begin{equation}
\begin{aligned}
    e_{t}^{acc} = \text{mean}\left(\frac{||\boldsymbol{s}_{t}^{acc}||^2}{||\hat{\boldsymbol{s}}_{t}^{vel}||^2+\lambda^{reg}}\right),
    \label{eq: ref2}
\end{aligned}
\end{equation}
where $\lambda^{reg}$ is a hyperparameter adjusts the sensitivity, $\boldsymbol{s}_{t}^{acc}$ is the simulated DOF accelerations of the humanoid, and $\hat{\boldsymbol{s}}_{t}^{vel}$ is the reference DOF velocities.

\subsection{High-level Task Rewards}
\subsubsection{Throwing}
In this task, the objective is to throw the ball to approach a certain height, grab the rebound, and keep on throwing the ball. The goal-related task reward can be simply defined as:
\begin{equation}
\begin{aligned}
    r_{t}^{throwing} = \text{exp}(-|h_{t}^{ball}-2.5|),
    \label{eq: throw}
\end{aligned}
\end{equation}
where $h_{t}^{ball}$ is the ball height.

\subsubsection{Heading}
This task aims to dribble the ball to approach the target position. The task observation $\boldsymbol{h}_{t}$ contains the target position. We simply define the task reward as:
\begin{equation}
\begin{aligned}
    r_{t}^{heading} = \text{exp}(-||\boldsymbol{x}_{t}^{ball}-\boldsymbol{x}_{t}^{target}||^2),
    \label{eq: Heading}
\end{aligned}
\end{equation}
where $\boldsymbol{x}_{t}^{ball}$ is the ball position while $\boldsymbol{x}_{t}^{target}$ is the target position.

\subsubsection{Circling}
In the circling task, the objective is for the humanoid to dribble the ball around the target position following a target radius. The task observation $\boldsymbol{h}_{t}$ contains the target position and radius. The task reward can be defined as:
\begin{equation}
\begin{aligned}
    r_{t}^{circling} =r_{t}^{v}*\text{exp}(-|d^{target} - ||\boldsymbol{x}_{t}^{ball}-\boldsymbol{x}_{t}^{center}||^2|),
    \label{eq: Circling}
\end{aligned}
\end{equation}
where $d^{target}$ is the target radius and $\boldsymbol{x}_{t}^{center}$ is the center point around which the humanoid is required to circle. $r_{t}^{v}$ is a 
speed constraint that prevents the ball from staying still, defined as:
\begin{equation}
\begin{aligned}
    r_{t}^{v} = \begin{cases}
    1,  & \text{if} \quad ||\boldsymbol{v}_t^{ball}||^2 >0.5 \\
    0.1, & \text{else} \\
    \end{cases}\\
\end{aligned}
\end{equation}
where $\boldsymbol{v}_t^{ball}$ is the ball velocity.
\subsubsection{Scoring}
To further validate our method's capability to combine a diverse set of skills for precise operations, we consider the scoring task. In this task, the objective is to shot the ball precisely into a randomly positioned basket. The task observation $\boldsymbol{h}_{t}$ contains the basket position. The reward function consists of four parts:
\begin{equation}
\begin{aligned}
    r_{t}^{scoring} = r_{t}^{v} *(r_{t}^{heading} + r_{t}^{bonus} + 0.2*r_{t}^{throwing}),
    \label{eq: Shot}
\end{aligned}
\end{equation}
where $r_{t}^{throwing}$ rewards the ball height, $r_{t}^{heading}$ encourages the ball to move close to the basket, $r_{t}^{v}$ prevents the ball from staying still, and $r_{t}^{bonus}$ is a bonus for a score, defined as:
\begin{equation}
\begin{aligned}
    r_{t}^{bonus} = \begin{cases}
    1,  & \text{if scored} \\
    0, & \text{else} \\
    \end{cases}\\
\end{aligned}
\end{equation}

\subsection{Success Rate}
\label{sec: success rate}
To evaluate the success rate of skills, we introduced a set of skill-specific rules to determine the success or failure of skill execution:
\begin{itemize}
    \item \textbf{Pickup}: When testing the pickup skill, we determine success by checking if the ball is lifted above 1 m after 10 seconds.
    \item \textbf{Dribble}: We have the humanoid dribble for 10 seconds, and if the root height of the humanoid is greater than 0.5 m and the distance between the ball and the humanoid root is less than 1.5 m, we consider the frame to be valid. The success rate is calculated as the proportion of valid frames to the total number of frames in 10 seconds.
    \item \textbf{Layup \& Shot}: We consider a success if the distance between the ball's maximum height and the target height is less than 0.1 m, and the body root height is above 0.5 m.
    \item \textbf{Throwing}: We evaluate success by checking if the ball remains above 0.3 m within 10 seconds after the first throw.
    \item \textbf{Heading}: We determine success if the distance between the ball and the target position is less than 0.5 m.
    \item \textbf{Scoring}: We consider a success if the ball's maximum height is above 2.5 m, the distance between the ball and target position is less than 0.3 m, and there is no contact between the ball and the humanoid. 
    \item \textbf{Circling}: A frame is considered valid if the distance between the ball and the target point differs from the set radius by less than 0.5 m and the ball's speed exceeds 0.5 m/s. The success rate is calculated as the proportion of valid frames to the total number of frames in the run.
\end{itemize}

\subsection{Hyperparameters}
\label{app: hyperparameters}
The hyperparameter configurations employed during the pre-training phase of the skill policy are detailed in Tab.~\ref{tab: params_llc}, while the hyperparameters utilized for the training of the high-level controller are presented in Tab.~\ref{tab: params_hlc}. Additionally, Tab.~\ref{tab: params_rewards} displays the hyperparameter settings for all sub-rewards involved.

\subsection{Details of Compared Methods}
\subsubsection{Variants in Skill Learning}
Since our approach is the first to learn interaction skills from demonstration, there are no direct benchmarks for comparison. Therefore, we adapt reward strategies commonly used in locomotion imitation, i.e., DeepMimic~\cite{DeepMimic} and AMP~\cite{amp}, to an object-inclusive setting. 
Specifically, we adapt our reward to the styles of DeepMimic~\cite{DeepMimic} and AMP~\cite{amp} while keeping the other components unchanged for a fair comparison. 
We denote these variant versions as DeepMimic* and AMP*. We will next delineate the specific differences in the reward functions of these variants. The sub-rewards below are consistent to that defined in Sec.~\ref{app: Kin Reward}. Unless specified, the hyperparameters of these sub-rewards are the same, as shown in Tab.~\ref{tab: params_rewards}.
\paragraph{DeepMimic*}
The reward function is
\begin{equation}
\begin{aligned}
    r_{t} = r_{t}^{p} + r_{t}^{r} + r_{t}^{rv} +  r_{t}^{o}.
    \label{eq: -dm}
\end{aligned}
\end{equation}
The hyperparameters of these sub-rewards are shown in Tab.~\ref{tab: params_rewards}. Unlike DeepMimic \cite{DeepMimic}, we do not incorporate phase information as policy input. This decision is made because phase-based methods face several practical limitations: they cannot operate continuously when reference data fails to form a complete cycle, and they show limited resistance to interference. All methods presented in this paper deliberately avoid using phase information.

\paragraph{AMP*}
The reward function is
\begin{equation}
    r_t = - \text{log} \left(1 - D(\boldsymbol{s}_t, \boldsymbol{s}_{t+1})\right),
    \label{eqn: amp*}
\end{equation}
where $\boldsymbol{s}$ represents the HOI state which includes body and object state. $D$ denotes the discriminator. 
\paragraph{SkillMimic w/o Multiplication}
The reward function is
\begin{equation}
\begin{aligned}
    r_{t} = r_{t}^{b}+r_{t}^{o}+r_{t}^{rel}+r_{t}^{reg}+r_{t}^{cg}.
    \label{eq: -m}
\end{aligned}
\end{equation}
\paragraph{SkillMimic w/o CGR}
The reward function is
\begin{equation}
\begin{aligned}
    r_{t} = r_{t}^{b}*r_{t}^{o}*r_{t}^{rel}*r_{t}^{reg}.
    \label{eq: -cgr}
\end{aligned}
\end{equation}


\begin{table}[t]
\centering  
\resizebox{0.9\linewidth}{!}{
\begin{tabular}{|l|c|}
\hline
{\bf Parameter} & {\bf Value}  \\ \hline
    $\mathrm{dim}(\boldsymbol{c})$ Skill Embedding Dimension &  $64$  \\ \hline
    $\Sigma_\pi$ Action Distribution Variance &  $0.055$  \\ \hline
    Samples Per Update Iteration &  $65536$  \\ \hline
    Policy/Value Function Minibatch Size &  $16384$  \\ \hline
    $\gamma$ Discount &  $0.99$  \\ \hline
    Adam Stepsize & $2 \times 10^{-5}$ \\ \hline
    GAE($\lambda$) &  $0.95$  \\ \hline
    TD($\lambda$) &  $0.95$  \\ \hline
    PPO Clip Threshold &  $0.2$  \\ \hline
    $T$ Episode Length &  $60$  \\ \hline
\end{tabular}
}
\caption{Hyperparameters for training skill policy.}
\label{tab: params_llc}
\end{table}

\begin{table}[t]
\centering  
\resizebox{0.9\linewidth}{!}{
\begin{tabular}{|l|c|}
\hline
{\bf Parameter} & {\bf Value}  \\ \hline
    $\Sigma_\pi$ Action Distribution Variance &  $0.055$  \\ \hline
    Samples Per Update Iteration &  $65536$  \\ \hline
    Policy/Value Function Minibatch Size &  $16384$  \\ \hline
    $\gamma$ Discount &  $0.99$  \\ \hline
    Adam Stepsize & $2 \times 10^{-5}$ \\ \hline
    GAE($\lambda$) &  $0.95$  \\ \hline
    TD($\lambda$) &  $0.95$  \\ \hline
    PPO Clip Threshold &  $0.2$  \\ \hline
    $T$ Episode Length &  $800$  \\ \hline
\end{tabular}
}
\caption{Hyperparameters for training high-level controller.}
\label{tab: params_hlc}
\end{table}

\begin{table}[t]
\centering  
\resizebox{1.\linewidth}{!}{
\begin{tabular}{|l|c|c|}
\hline
{\bf Parameter} & {\bf SM} & {\bf DM*}  \\ \hline
    $\lambda^{p}$ Sensitivity of Key Body Position Error &  $20$ &  $20$  \\ \hline
    $\lambda^{r}$ Sensitivity of DOF Rotation Error &  $20$ &  $2$  \\ \hline
    $\lambda^{pv}$ Sensitivity of Key Body Velocity Error &  $0$  &  $-$ \\ \hline
    $\lambda^{rv}$ Sensitivity of DOF Rotation Velocity Error &  $0$ &  $0.1$  \\ \hline
    $\lambda^{op}$ Sensitivity of Object Position Error & $20$ &  $20$ \\ \hline
    $\lambda^{or}$ Sensitivity of Object Rotation Error &  $0$  &  $-$ \\ \hline
    $\lambda^{opv}$ Sensitivity of Object Velocity Error &  $0$ &  $-$  \\ \hline
    $\lambda^{orv}$ Sensitivity of Object Angular Velocity Error &  $0$ &  $-$  \\ \hline
    $\lambda^{rel}$ Sensitivity of Relative Position Error &  $20$ &  $-$  \\ \hline
    $\boldsymbol{\lambda^{cg}}[0]$ Sensitivity of Ball-Hands Contact Error &  $5$ &  $-$  \\ \hline
    $\boldsymbol{\lambda^{cg}}[1]$ Sensitivity of Ball-Body Contact Error &  $5$ &  $-$  \\ \hline
    $\boldsymbol{\lambda^{cg}}[1]$ Sensitivity of Body-Hands Contact Error &  $5$ &  $-$  \\ \hline
     $\lambda^{reg}$ Sensitivity of Velocity Regularization &  $10^{-12}$ &  $-$  \\ \hline
\end{tabular}
}
\caption{Hyperparameters of Sub-Rewards. SM denotes SkillMimic, and DM* denotes SkillMimic with DeepMimic-style rewards.}
\label{tab: params_rewards}
\end{table}

\subsubsection{Variants in High-Level Tasks}
To evaluate the performance of our method on high-level tasks, we established three sets of experiments for comparison: one involving training from scratch and the other two utilizing ASE \cite{ase}, which first train a Low-Level Controller (LLC) using GAIL \cite{ho2016generative} then train a high-level controller to control the LLC. For fair comparison, we construct two versions of LLC for ASE. The first LLC follows the original style of ASE \cite{ase} which learns locomotion skills without considering object motion. We denote this method as ASE in our experiment. The second LLC follows the style of SkillMimic which learns interaction skills and considers object motion. We denote this object-inclusive LLC as ASE* in our experiment. The reward of both LLCs share the same formulation:
\begin{equation}
    r_t = - \text{log} \left(1 - D(\boldsymbol{s}_t, \boldsymbol{s}_{t+1})\right) + \beta \ \text{log} \ q\left(\boldsymbol{z}_t | \boldsymbol{s}_t, \boldsymbol{s}_{t+1}\right),
    \label{eqn: ase}
\end{equation}
The difference is in the representations of $\boldsymbol{s}$. For ASE*, $\boldsymbol{s}$ contains the body and object state. For ASE, $\boldsymbol{s}$ represents the body-only state (object is not considered). $D$ denotes the discriminator, $q$ denotes the encoder and $\boldsymbol{z}$ represents the latent code.

Subsequently, we trained High-Level Controllers (HLC) to reuse these pre-trained LLCs for high-level tasks. For fair comparison, the task rewards are the same and the network size of the HLC is identical to that used in our approach. The HLC of ASE and ASE* outputs continuous latent variables, whereas our HLC outputs discrete skill conditions.

\end{document}